\documentclass[10pt,twocolumn,letterpaper]{article}

\usepackage[pagenumbers]{cvpr} 

\definecolor{cvprblue}{rgb}{0.21,0.49,0.74}
\usepackage[pagebackref,breaklinks,colorlinks,allcolors=cvprblue]{hyperref}
\usepackage[accsupp]{axessibility}
\usepackage{adjustbox}

\def\paperID{10146} 
\def\confName{CVPR}
\def\confYear{2025}

\title{3D-GSW: 3D Gaussian Splatting for Robust Watermarking}

\author{
  Youngdong Jang\textsuperscript{\normalfont 1} \quad Hyunje Park \textsuperscript{\normalfont 1} \quad
  Feng Yang \textsuperscript{\normalfont 2} \quad Heeju Ko \textsuperscript{\normalfont 1} \quad Euijin Choo\textsuperscript{\normalfont 3} \quad Sangpil Kim\textsuperscript{\normalfont 1}\thanks{Corresponding Author} \\
  \\
  \textsuperscript{\normalfont 1} Korea University \quad \textsuperscript{\normalfont 2} Google DeepMind \quad \textsuperscript{\normalfont 3} University of Alberta
}

\begin{document}
\maketitle
\begin{abstract} 
As 3D Gaussian Splatting~(3D-GS) gains significant attention and its commercial usage increases, the need for watermarking technologies to prevent unauthorized use of the 3D-GS models and rendered images has become increasingly important.
In this paper, we introduce a robust watermarking method for 3D-GS that secures copyright of both the model and its rendered images.
Our proposed method remains robust against distortions in rendered images and model attacks while maintaining high rendering quality. 
To achieve these objectives, we present Frequency-Guided Densification~(FGD), which removes 3D Gaussians based on their contribution to rendering quality, enhancing real-time rendering and the robustness of the message. FGD utilizes Discrete Fourier Transform to split 3D Gaussians in high-frequency areas, improving rendering quality. 
Furthermore, we employ a gradient mask for 3D Gaussians and design a wavelet-subband loss to enhance rendering quality. Our experiments show that our method embeds the message in the rendered images invisibly and robustly against various attacks, including model distortion. Our method achieves superior performance in both rendering quality and watermark robustness while improving real-time rendering efficiency. Project page: \url{https://kuai-lab.github.io/cvpr20253dgsw/}  
\end{abstract}
\vspace{-1em}

\section{Introduction}
\label{sec:intro}

3D representation has been at the center of computer vision and graphics. Such technology plays a pivotal role in various applications and industries, \textit{e.g.}, movies, games, and the Metaverse industry. Since Neural Radiance Field~\cite{mildenhall2021nerf}~(NeRF) has shown great success in 3D representation due to photo-realistic rendering quality, it has been at the forefront of 3D content creation.
\begin{figure}[htb!]
    \begin{center}
        \includegraphics[width=0.9\linewidth]{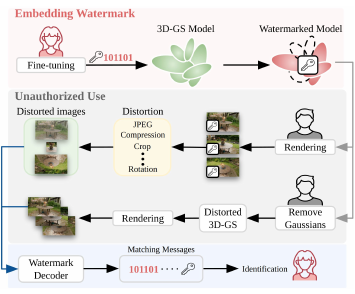}
    \end{center}
    \vspace{-2em}
    \caption{The unauthorized use of the 3D Gaussian Splatting model. Our method ensures that the watermark remains detectable even in distorted images and under model attacks.} 
    \label{fig:flow}
    \vspace{-1.7em}
\end{figure}

Recently, 3D Gaussian Splatting~\cite{kerbl20233d} (3D-GS) has gained attention for its real-time rendering performance and high rendering quality, compared to other radiance field methods~\cite{chen2022tensorf, fridovich2022plenoxels, mildenhall2021nerf, muller2022instant}. 3D-GS is an explicit representation that uses trainable 3D Gaussians. 
This explicit property enhances the capability of 3D-GS to generate 3D assets. Due to these properties, 3D-GS has been a transformative 3D representation.  

While 3D-GS has advanced and practical usage has increased, it raises concerns about the unauthorized use of its 3D assets. Therefore, attempts have been made to develop digital watermarking for radiance fields to address this problem, such as WateRF~\cite{jang2024waterf} , which integrates watermark embedding into the rendering process. However, this method presents several challenges when applied directly to 3D-GS. First, achieving high-fidelity rendering requires redundant 3D Gaussians, which leads to substantial memory and storage overheads, especially for large-scale scenes. This also makes embedding a large amount of watermark computationally expensive, significantly increasing processing time. 
Secondly, there are many 3D Gaussians that have minimal impact on the rendered image. Their minimal impact on the rendered image makes it difficult to embed the watermark robustly.

To address these issues, we propose Frequency-Guided Densification~(FGD) to reduce the number of 3D Gaussians to ensure both real-time rendering and robust message embedding. FGD consists of two phases. In the first phase, we remove 3D Gaussians based on their contribution to the rendering quality. The remaining 3D Gaussians, which significantly impact the rendered image, enable robust message embedding.
In the second phase, we utilize two properties to enhance rendering quality: 1) Smaller 3D Gaussians have minimal impact on the rendered image~\cite{lee2023compact}. 2) The human visual system is less sensitive to high-frequency areas~\cite{lee2012perceptual}. To identify high-frequency areas, we apply Discrete Fourier Transform (DFT) to the rendered image in a patch-wise manner and measure the intensity of high-frequency. After that, 3D Gaussians with strong high-frequency intensity are split into smaller ones to ensure the rendering quality. 

Furthermore, significant changes in the parameters of 3D-GS, optimized for high rendering quality, lead to substantial variations in the rendered output.
To minimize the adjustments, we utilize a gradient mask derived from the pre-trained parameters, transmitting smaller gradients to 3D-GS during optimization. In this way, the rendering quality is not significantly decreased.
To further enhance rendering quality, we design a wavelet-subband loss. Since we split 3D Gaussians in high-frequency areas, the wavelet-subband loss enhances the local structure by leveraging only high-frequency components.

Our experimental results show that our method effectively fine-tunes 3D-GS to embed the watermark into the rendered images from all viewpoints. We also evaluate the robustness of our method under various attacks, including image distortion and model attacks. We compare the performance of our method with other methods~\cite{li2023steganerf, jang2024waterf} and demonstrate that our method outperforms other state-of-the-art radiance field watermarking methods across all metrics. Our main contributions are summarized as follows: 
\begin{itemize}
    \item We propose frequency-guided densification to effectively remove 3D Gaussians without compromising the rendering quality, to embed a robust message in the rendered image while enhancing the real-time rendering.
    \item We propose a gradient mask mechanism that minimizes gradients to preserve similarity to the pre-trained 3D-GS and maintain high rendering quality.
    \item We introduce a wavelet-subband loss to enhance rendering quality, particularly in high-frequency areas.
    \item The proposed method achieves superior performance and demonstrates robustness against various types of attacks, including both image and model distortions. 
\end{itemize}
\section{Related work}
\label{sec:formatting}

\subsection{3D Gaussian Splatting}
Recently, 3D Gaussian Splatting (3D-GS)~\cite{kerbl20233d} has brought a paradigm shift in the radiance field by introducing an explicit representation and differentiable point-based splatting methods, allowing for real-time rendering. 3D-GS has been applied to various research areas, including 3D reconstruction~\cite{lu2024scaffold,lin2024vastgaussian,jiang2024gaussianshader,yan2024multi}, dynamic scenes~\cite{lu20243d, yang2024deformable, wu20244d, huang2024sc}, avatar~\cite{shao2024splattingavatar, li2024animatable, lei2024gart,moreau2024human} and generation~\cite{chen2024text, liang2024luciddreamer, liu2024humangaussian, ling2024align}. Its capability and efficiency have made 3D-GS widely used, positioning it at the forefront of 3D asset generation. As adoption grows across various applications, ensuring the integrity and reliability of generated content has become increasingly important. Therefore, the copyright protection of 3D-GS-generated assets has become essential aspect.

\subsection{Frequency Transform}

\noindent \textbf{Discrete Fourier Transform (DFT)} has played a crucial role in signal processing and image processing. Recent research~\cite{jiang2021focal,rao2021global,jia2023fourier,yu2022deep} has applied DFT to images and leveraged frequency signals to improve model performance and analyze images. Baig~\cite{baig2022dft} utilizes DFT to estimate the quality of blurred images globally. Rao~\cite{radha2023deep} leverages this ability of DFT to acquire global information about images. According to these studies, DFT can efficiently analyze global information in images. Since we need to analyze global frequency signal strength across image patches, we use DFT to transform the rendered images in a patch-wise manner.
\\
\vspace{-0.7em}

\noindent \textbf{Discrete Wavelet Transform (DWT)} analyzes signals or images by decomposing them into components with different frequencies and resolutions. DWT is particularly effective at capturing local information. In the previous works~\cite{tuba2021image, mohideen2008image, pimpalkhute2021digital}, DWT has been applied to images for denoising. Tian~\cite{tian2023multi} utilizes DWT as it provides both spatial and frequency information through multi-resolution analysis, enabling effective noise suppression and detailed image restoration. For the radiance field, previous works~\cite{xu2023wavenerf, jang2024waterf, rho2023masked, lou2024darenerf} show the compatibility between the radiance field and DWT. Leveraging these advantages, we utilize DWT to compute loss functions between high-frequency local information, thereby enhancing rendering quality.

 \begin{figure*}[ht!]
    \begin{center}
        \includegraphics[width=1\textwidth ]{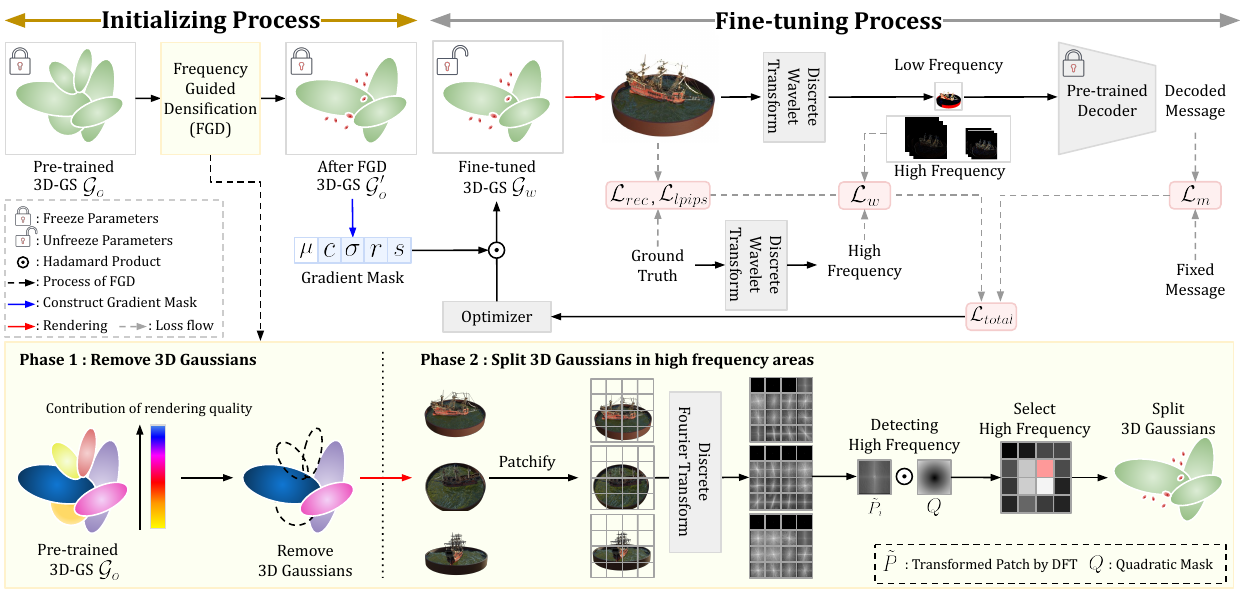}
    \end{center}
    \vspace{-1.5em}
    \caption{\textbf{3D-GSW Overview}. Before fine-tuning 3D-GS, Frequency-Guided Densification (FGD) removes 3D Gaussians based on their contribution to the rendering quality and splits 3D Gaussians in high-frequency areas into smaller ones. We also construct a gradient mask based on the parameters of an FGD-processed 3D-GS. During the fine-tuning, we apply the Discrete Wavelet Transform~(DWT) to the rendered image for robustness, using the low frequency as input to a pre-trained message decoder. For rendering quality, we design a wavelet-subbands loss that utilizes only high-frequency subbands. Finally, 3D-GS is optimized through the $\mathcal{L}_{total}$.}
    \vspace{-1em}
    \label{fig:overview}
\end{figure*}

\subsection{Steganography and Digital Watermarking}
\noindent \textbf{Steganography} is employed to maintain the confidentiality of information by embedding it invisibly within digital assets. Recently, there has been growing interest in applying steganography to the radiance field~\cite{li2023steganerf,biswal2024steganerv,dong2024stega4nerf}. StegaNeRF~\cite{li2023steganerf} fine-tunes the pre-trained radiance fields model to invisibly embed images into the rendered image. For 3D-GS, GS-hider~\cite{zhang2024gs} invisibly embeds 3D scenes and images into point clouds.
\\
\vspace{-0.7em}

\noindent \textbf{Digital watermarking} protects digital assets by identifying the copyrights. The main difference lies in the priority of data embedding. The primary goal of digital watermarking is robustness, ensuring that embedded data can be detected even after distortions, whereas steganography prioritizes invisibility. 
To achieve robustness, the traditional watermarking methods~\cite{barni2001improved, shensa1992discrete, raval2003discrete, tao2004robust} have utilized DWT, embedding into the subbands of DWT. HiDDeN~\cite{zhu2018hidden} is the end-to-end deep learning watermarking method, which embeds the robust message by adding a noise layer. For radiance fields watermarking, CopyRNeRF~\cite{luo2023copyrnerf} explores embedding the message into the rendered image from implicit NeRF. WateRF~\cite{jang2024waterf} enhances both high rendering quality and robustness of watermarks through DWT. In this paper, we introduce a robust digital watermarking method for 3D-GS.
\section{Method}

\subsection{Preliminary}

3D-GS~\cite{kerbl20233d} represents the 3D world with a set of 3D Gaussian primitives, each defined as:
\vspace{-.2em}
\begin{equation}
    G(\mathbf{x} ; \mu, \Sigma) = e^{-\frac{1}{2}(\mathbf{x}-\mu)^{\mathrm{T}} \Sigma^{-1}(\mathbf{x}-\mu)}
\end{equation}
, where the parameters mean $\mu$ and covariance $\Sigma$ determine spatial distribution. To render these primitives onto an image plane, each 3D Gaussian is projected into 2D-pixel space and forms a 2D Gaussian primitive $\hat{G}$ by projective transform and its Jacobian evaluation $\mu$. The 2D 
 Gaussian primitives are depth-ordered, rasterized, and alpha-blended using transmittance $T_i$ as a weight to form an image:
\vspace{-.5em}
\begin{equation}
    \begin{aligned}
    I_{\pi}[x, y] = \sum_{i \in N_\mathcal{G}} c_i \alpha_i T_i, \; \text{where } T_i=\prod_{j=1}^{i-1}(1-\alpha_j) \;\; 
    \label{eq:render}
    \end{aligned}
\end{equation}
\begin{equation}
    \begin{aligned}
    \alpha_i = \sigma_i \hat{G}^{\pi}_i([x, y];\hat{\mu}, \hat{\Sigma})
    \label{eq:alpha}
    \end{aligned}
\end{equation}
, where $\pi$, $c_i$, $\sigma_i$, and $\alpha_i$ are the viewpoint, color, opacity, and density of each Gaussian primitive evaluated at each pixel. $N_\mathcal{G}$ denotes the set of depth-ordered 2D Gaussian primitives that are present in the selected viewpoint.

\subsection{Fine-tuning 3D Gaussian Splatting}  

As shown in Fig.~\ref{fig:overview}, we fine-tune the pre-trained 3D Gaussian Splatting (3D-GS) $\mathcal{G}_o$ into $\mathcal{G}_w$ to ensure the rendered images from all viewpoints contain a binary message $M = (m_1, ..., m_N) \in \{0,1\}^N$. To achieve this, we utilize a pre-trained message decoder, HiDDeN~\cite{zhu2018hidden}, denoted as $D_m$. Before fine-tuning, to enhance robustness, we employ Frequency-Guided Densification (FGD) to remove 3D Gaussians with minimal impact on the rendered image and split 3D Gaussians in high-frequency areas (see Sec.~\ref{contribution} and Sec.~\ref{Densification}). After that, we construct a gradient mask based on the FGD-processed 3D-GS $\mathcal{G}^{\prime}_o$ (see Sec.~\ref{Mask}) to ensure high rendering quality. 
In the fine-tuning process, $\mathcal{G}_w$ renders an image $I_w \in \mathbb{R}^{3 \times H \times W}$.
$I_w$ is transformed into the wavelet subbands $\{ LL_l, LH_l, HL_l, HH_l \}$
, where $l$ denotes the level of DWT. L and H are respectively denoted as low and high. Following the previous work~\cite{jang2024waterf}, we choose the $LL_2$ subband as input  $D_m$ and decode the message $M' = D_m(LL_2)$, ensuring efficient and robust message embedding. Additionally, we employ high-frequency subbands for the proposed wavelet-subband loss. Further details are provided in the following sections.

\subsection{Measure Contribution of Rendering Quality}
\label{contribution}

The pre-trained 3D-GS includes redundant 3D Gaussians to ensure high-quality rendering. 
Because 3D Gaussians with minimal impact on rendering quality can also carry the message, it tends to be weakly embedded in the rendered image.
To address this limitation, we remove 3D Gaussians with minimal impact on the rendered image before the fine-tuning process. 
Inspired by error-based densification~\cite{bulò2024revising}, we measure the contribution of each 3D Gaussian to the rendering quality using the auxiliary loss function $L_\pi^{aux}$ with a new color parameter set $C'$ for the viewpoint $\pi$:
\vspace{-.5em}
\begin{equation}
    L_\pi^{aux} := \frac{\sum_{x,y \in Pix} \mathcal{E}_\pi[x,y] I^{c'}_\pi[x,y]}{H \times W}
\end{equation}
\begin{equation}
   \mathcal{E}_\pi = \vert I_\pi^{c'} - I_\pi^{gt} \vert
\end{equation}
, where $I^{c'}_\pi \in \mathbb{R}^{3 \times H \times W}$ and $I_\pi^{gt} \in \mathbb{R}^{3 \times H \times W}$ are respectively denoted as a rendered image with $C'$ and ground truth. We replace the parameters $C$ with $C'$ only when $\mathcal{G}_o$ renders $I^{c'}_\pi$, and set all of its values to zeros. During the backward process, the gradients of the auxiliary loss with respect to ${C}'$ are derived as follows: 
\vspace{-.5em}
\begin{equation}
    V_\pi = \frac{{\partial{L_\pi^{aux}}}}{{\partial{C}'}} = \sum_{x,y \in Pix} \mathcal{E}_\pi[x,y] w_{\pi} 
\end{equation}
\begin{equation}
    w_{\pi} = \sum_{i \in N_\mathcal{G}} c_i \alpha_i T_i
\end{equation}
, where $c_i$, $\alpha_i$ and $T_i$ are respectively denoted as the color, the density, and the transmittance of each 3D Gaussian. We utilize this $V_\pi$ $\in$ $\mathbb{R}^{N_\mathcal{G}\times{3}}$, as the contribution for a rendered quality at $\pi$, as it reflects each 3D Gaussian's contribution to the color of the rendered image. 

\subsection{Frequency-Guided Densification (FGD)}
\label{Densification}

Our method aims to embed the message $M$ robustly into the rendered image to ensure fast embedding and real-time rendering speed without a decrease in rendering quality. To achieve these objectives, we propose Frequency-Guided Densification (FGD), which removes 3D Gaussians, which have minimal impact on the rendered image, and splits 3D Gaussians in the high-frequency areas into smaller ones. 

FGD consists of two phases to achieve these goals. First, the pre-trained $\mathcal{G}_o$ renders the image $I_{\pi}^{c'}$ from all viewpoints, and we derive $V_{\pi}$ from the rendered images. Based on $V_{\pi}$, we remove 3D Gaussians that have negligible impact on the rendering quality. Second, since large scenes require substantial memory, images rendered by 3D-GS with 3D Gaussians removed are divided into patches $P \in \mathbb{R}^{3 \times M \times N}$ to improve memory efficiency. Since FGD identifies patches with strong high-frequency signals, we utilize the Discrete Fourier Transform~(DFT) for the global frequency analysis. The DFT is defined as follows:
\vspace{-0.5em}
\begin{equation}
   F[u,v] = \sum_{m=0}^{M-1} \sum_{n=0}^{N-1} f[m,n] e^{-j2\pi \left(\frac{u}{M}m + \frac{v}{N}n\right)}
    \label{eq:dft}
\end{equation}
, where $f$ and $F$ are respectively denoted as spatial-domain pixel value at spatial-domain image coordinate $(m,n)$ and frequency-domain pixel value at the frequency-domain image coordinate $(u,v)$.
We transform the spatial-domain patch $P$ into the frequency domain using DFT, revealing a complete spectrum of frequency components, i.e., $\tilde{P} = \mathbb{R}(F(P)) \in \mathbb{R}^{3 \times U \times V}$. The transformed patch $\tilde{P}$ undergoes Hadamard product $\odot$ with a mask $Q \in \mathbb{R}^{3 \times U \times V}$, designed to emphasize high-frequency signals, and the intensity of high-frequency $E$ is computed as follows:
\vspace{-.7em}
\begin{equation}
        Q[u, v] = (\frac{2u-U}{U})^2 + (\frac{2v-V}{V})^2
    \label{eq:patch mask}
\end{equation}
\begin{equation}
        E = \frac{\sum_{u, v} (\tilde{P} \odot Q)_{uv}}{U \times V}
    \label{eq:intensity of high-freq}
\end{equation}
, where $(u, v) \in \mathbb{R}^{U \times V}$. We select the top $K \%$ patch $\tilde{P}$ based on $E$ and track 3D Gaussians from the chosen patches. Based on $V_\pi$, we choose the 3D Gaussians that have less impact on the image and split them into smaller ones to enhance rendering quality. Therefore, we effectively reduce the number of 3D Gaussians to enhance rendering speed and maintain high rendering quality. 
With intensive optimization of 3D Gaussians that significantly impact rendering quality, a robust message can be embedded.

\begin{figure*}[htb!]
    \begin{center}
        \includegraphics[width=1\textwidth ]{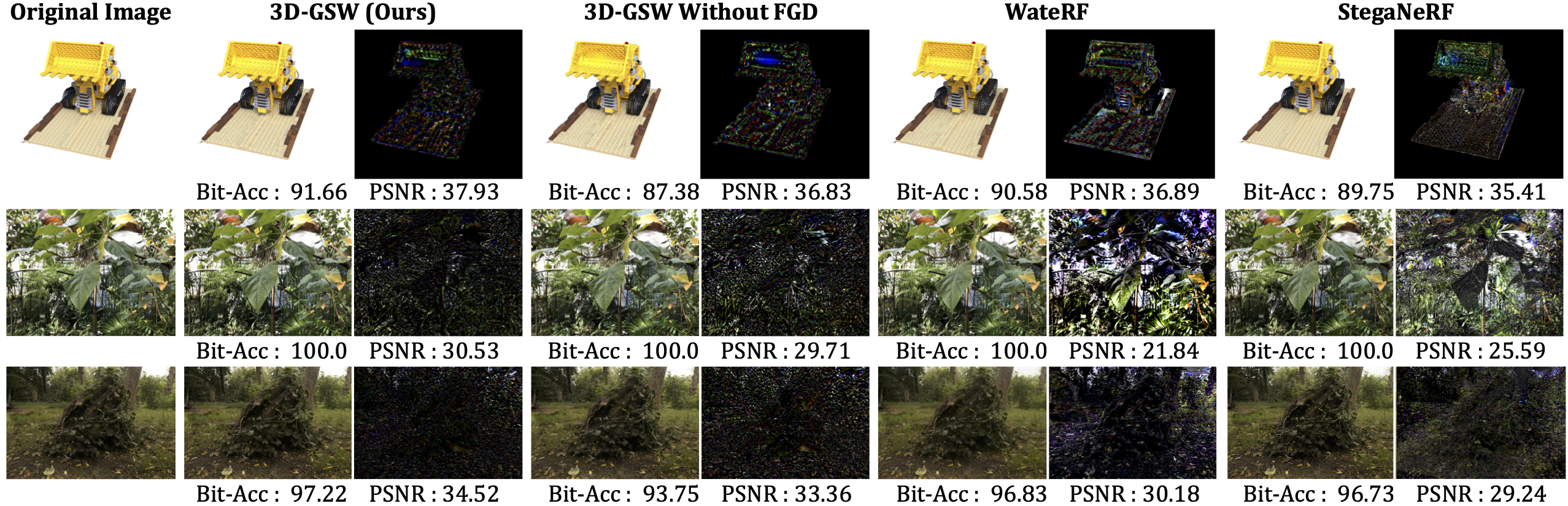}
    \end{center}
    \vspace{-1.5em}
    \caption{Rendering quality comparison of each baseline with our method. We doubled the scale of the difference map. Our method outperforms others in bit accuracy and rendering quality, using 32-bit messages for the qualitative results.}
    \vspace{-1em}
    \label{fig:figure3}
\end{figure*}

\subsection{Gradient Mask for 3D  Gaussian Splatting}
\label{Mask}
Since 3D-GS $\mathcal{G}^{\prime}_o$ passed through FGD renders high-quality images, we must embed the message without compromising rendering quality. To achieve this, we further reduce the gradient magnitude during fine-tuning to minimize changes in the parameters $\theta$ of $\mathcal{G}^{\prime}_o$. The parameters $\theta$ consist of position $\mu$, color $c$, opacity $\sigma$, rotation $r$, and scale $s$.

While StegaNeRF~\cite{li2023steganerf} uses a gradient mask to modify the gradient, applying this method to 3D-GS is challenging due to the zero values in its parameters.
To avoid dividing by zero and further reduce the magnitude of the gradient to minimize changes in parameters, we incorporate an exponential function into the mask calculation.
To reduce the gradient size of the parameter $\theta$ for each 3D Gaussian, the gradient mask $z \in \mathbb{R}^{N_{\mathcal{G}^{\prime}_o}}$ is calculated as follows :
\begin{equation}
\begin{aligned}
w = \frac{1}{e^{\vert \theta \vert^{\beta}}}, \quad  z = \frac{w}{\sum_{i=1}^{N_{\mathcal{G}^{\prime}_o}} w_i} 
\end{aligned}
\label{eq:adaptive mask}
\end{equation}
, where i and $\beta > 0$ are respectively denoted as the index of 3D Gaussians and the strength of gradient manipulation.  We calculate the mask $z$ for each parameter, $c$,  $\sigma$,  $r$ and  $s$. The gradient of the positions parameter, in particular $\mu$, remains close to zero. Therefore, we apply a gradient mask to the parameters, except for $\mu$. During the fine-tuning, the gradient is masked as $\frac{\partial \mathcal{L}_{total}} {\partial \theta} \odot z$ , where $\mathcal{L}_{total}$ is Eq.~\ref{eq:toal_loss} and $\odot$ denotes Hadamard product. Since small gradients are transmitted to 3D-GS, our gradient mask enables message embedding while preserving high rendering quality.

\subsection{Losses}
\label{Loss}

We model the objective of 3D-GS watermarking by optimizing: 1) the reconstruction loss, 2) the LPIPS loss~\cite{zhang2018perceptual} 3) the wavelet-subband loss, and 3) the message loss.
For the reconstruction loss,  $\mathcal{L}_{rec}$, we measure the difference between the original image $I_o$ and the watermarked image $I_w$. We employ the loss function $\mathcal{L}_1$:
\begin{equation}
        \mathcal{L}_{rec} = \mathcal{L}_1(I_w,I_o)
    \label{eq:rec loss}
\end{equation}

For the LPIPS loss, $\mathcal{L}_{lpips}$, we evaluate the perceptual similarity between the feature maps of $I_o$ and $I_w$. This loss is typically computed by extracting feature maps from a pre-trained network $f(x)$:
\begin{equation}
    \mathcal{L}_{lpips} = \sum_{l} \omega_l \cdot \mathbb{E} \left[ ( f(I_w) - f(I_o) )^2 \right]
\end{equation}
, where \( l \) and \( \omega_l \) are respectively denoted as the layer index of the pre-trained network and the learned scaling factor.

Since we modify 3D Gaussians in the high-frequency areas, we design a wavelet-subband loss $\mathcal{L}_{w}$ to further enhance the rendering quality of high-frequency areas. Since DWT effectively analyzes local details using several subbands, we only employ high-frequency subbands $\{LH_l, HL_l, HH_l\}$ to improve the rendering quality during embedding of the message. To utilize $\mathcal{L}_{w}$, $I_o$ is transformed into wavelet subbands $\{LL^{gt}_l, LH^{gt}_l, HL^{gt}_l, HH^{gt}_l\}$. We employ the loss function $\mathcal{L}_1$:
\begin{equation}
    \mathcal{L}_w = \sum_{l} \sum_{S} \mathcal{L}_1 (S_l, S_l^{gt}),
    \; \text{where } S \in \{LH, HL, HH\}
    \label{eq:dwt_loss}
\end{equation}

For the message loss $\mathcal{L}_{m}$, we employ a sigmoid function to confine the extracted message $M'$ within the range of [0, 1]. The message loss is a Binary Cross Entropy between the fixed message $M$ and the sigmoid $sg(M')$:
\begin{equation}
    \mathcal{L}_{m} = - \sum_{i=1}^N ( M_i \cdot \log sg(M'_i) + (1-M_i) \cdot \log(1-sg(M'_i)))
    \label{eq:message_loss}
\end{equation}

Finally, 3D-GS is optimized with the total loss, which is the weighted sum of all losses:
\begin{equation}
        \mathcal{L}_{total} = \lambda_{rec} \mathcal{L}_{rec} + \lambda_{lpips} \mathcal{L}_{lpips} + \lambda_{w} \mathcal{L}_{w} + \lambda_{m} \mathcal{L}_{m}
    \label{eq:toal_loss}
\end{equation}

\section{Experiments}

\begin{figure*}[htb!]
    \begin{center}
        \includegraphics[width=1\textwidth ]{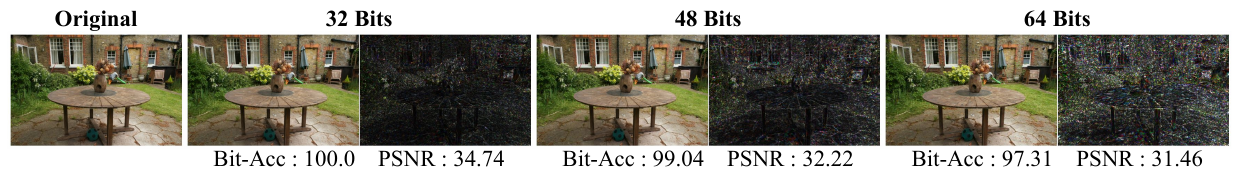}
    \end{center}
    \vspace{-2em}
    \caption{We present a rendering quality comparison for 32-bit, 48-bit, and 64-bit messages. The differences ($\times$2) between the watermarked image and the original image. Since manipulated areas are high-frequency areas where the people's eyes are less sensitive, the rendered image with our method looks more realistic and natural.}
    \vspace{-1em}
    \label{fig:figure_length}
\end{figure*}

\begin{table*}[htb!]
\setlength{\tabcolsep}{3pt}
\centering
\footnotesize{
\begin{tabular}{@{}l|ccccc|ccccc|ccccc@{}}
\toprule
  & \multicolumn{4}{c}{32 bits} & & \multicolumn{4}{c}{48 bits} & & \multicolumn{4}{c}{64 bits} \\
   \cmidrule{2-5} \cmidrule{7-10} \cmidrule{12-15} 
    Methods   & Bit Acc↑ & PSNR ↑ & SSIM ↑ & LPIPS ↓ & & Bit Acc↑ & PSNR ↑ & SSIM ↑ & LPIPS ↓ & & Bit Acc↑ & PSNR ↑ & SSIM ↑ & LPIPS ↓  \\ \midrule
     StegaNeRF~\cite{li2023steganerf}+3D-GS~\cite{kerbl20233d} &  93.15 &   32.68 &  0.953  &   0.049 &&  89.43 &   32.72 &  0.954  & 0.048 && 85.27  & 30.66 & 0.925 & 0.092 \\
     WateRF~\cite{jang2024waterf}+3D-GS~\cite{kerbl20233d}  & 93.42 & 30.49 & 0.956 & 0.050 &&  84.16  &  29.92  & 0.951 & 0.053  &&  75.10 & 25.81 & 0.883 & 0.108 \\ \midrule
      3D-GSW without FGD  &  94.60  &  34.27  & 0.975 &  0.047 &&  86.69  &   30.46 & 0.896 & 0.074 && 82.49  & 28.22 & 0.893 & 0.077 \\
     3D-GSW (Ours)  & \textbf{97.37}   &  \textbf{35.08}  & \textbf{0.978} & \textbf{0.043}  &&  \textbf{93.72}  &  \textbf{33.31}  & \textbf{0.970} & \textbf{0.045} &&  \textbf{90.45} & \textbf{32.47} & \textbf{0.967} & \textbf{0.049} \\
\bottomrule
\end{tabular}}
\vspace{-1em}
\caption{Bit accuracy and quantitative comparison of rendering quality with baselines. We show the results in 32, 48, and 64 bits. The results are the average of Blender, LLFF, and Mip-NeRF 360 datasets. The best performances are highlighted in \textbf{bold}.}
\vspace{-1em}
\label{tab:invisibility_capacity}
\end{table*}


\subsection{Experimental Setting}
\noindent \textbf{Dataset \& Pre-trained 3D-GS.} We use Blender~\cite{mildenhall2021nerf}, LLFF~\cite{mildenhall2019local} and Mip-NeRF 360~\cite{barron2022mip}, which are considered standard in NeRF~\cite{mildenhall2021nerf} and 3D-GS~\cite{kerbl20233d} literature. We follow the conventional NeRF~\cite{mildenhall2021nerf} and 3D-GS~\cite{kerbl20233d}, wherein we compare the results using 25 scenes from the full Blender, LLFF, and Mip-NeRF 360 datasets.
\\
\vspace{-0.8em}

\noindent \textbf{Baseline.}
We compare our method (3D-GSW) with three strategies for fairness: 1) StegaNeRF~\cite{jang2024waterf}: the steganography method for NeRF models, which embed an image into the rendered image. We add three linear layers to the watermark decoder to enable message decoding. Additionally, to apply the mask of StegaNeRF~\cite{jang2024waterf}, we set the parameters of 3D-GS to a small value of zero to a small value $10^{-4}$. 2) WateRF~\cite{jang2024waterf} + 3D-GS~\cite{kerbl20233d}: currently the state-of-the-art watermarking method for NeRF models. 3) 3D-GSW without FGD: changing our method by removing FGD.
\\
\vspace{-0.8em}

\noindent \textbf{Implementation Details.} Our method is trained on a single A100 GPU. The training is completed with epochs ranging from 2 to 10. The iteration per epoch is the number of train viewpoints in the datasets. We use Adam~\cite{kingma2014adam} to optimize 3D-GS. For the decoder, we pre-train HiDDeN~\cite{zhu2018hidden} decoder for bits = $\{32,48,64\}$ and freeze the parameters during our fine-tuning process. We set $\lambda_{rec}$ = 1, $\lambda_{lpips}$ = 0.2, $\lambda_{w}$ = 0.3, and $\lambda_{m}$ = 0.4 in our experiments. We remove 3D Gaussians under $V_{\pi} = 10^{-8}$. Also, we set the patch size $|P|$ = 16, the K = 1\%, and $\beta$ = 4. Our experiments are conducted on five different seeds.
\\
\vspace{-0.8em}

\noindent \textbf{Evaluation.} We consider three important aspects of watermarks: 
1) \textbf{Invisibility}: We evaluate invisibility by using the Peak Signal-to-Noise Ratio (PSNR), Structural Similarity Index Measure (SSIM), and Learned Perceptual Image Patch Similarity (LPIPS)~\cite{zhang2018perceptual}. 
2) \textbf{Robustness}: We investigate robustness by measuring bit accuracy under various distortions. The following distortions for message extraction are considered: Gaussian noise~($\sigma$ = 0.1), Rotation~(random select between $ + \pi / 6$ and $ - \pi / 6$), Scaling~(75 $\%$ of the original), Gaussian blur~($\sigma$ = 0.1), Crop~(40 $\%$ of the original), JPEG compression~(50 $\%$ of the original), a combination of Gaussian Noise, Crop, JPEG Compression. Furthermore, we consider a distortion of the core model, such as removing 3D Gaussians~(20 \%), cloning 3D Gaussians~(20 \%) and adding Gaussian noise ($\sigma$ = 0.1) to the parameters of 3D-GS.
3) \textbf{Capacity}: We explore the bit accuracy across various message lengths, which are denoted as $M_b \in \{32,48,64\}$.

\begin{table*}[ht!]
\setlength{\tabcolsep}{3pt}
\begin{adjustbox}{max width=\textwidth,center}
{\large
\begin{tabular}{l|ccccccccc}
\toprule
\multicolumn{1}{c|}{} & \multicolumn{8}{c}{Bit Accuracy(\%) $\uparrow$} \\
\cmidrule{2-9}
\multicolumn{1}{c|}{Methods} & No Distortion & \begin{tabular}[c]{@{}c@{}}Gaussian Noise\\ ($\sigma$ = 0.1)\end{tabular} & \begin{tabular}[c]{c}Rotation\\ ($\pm \pi/6$)\end{tabular} & \begin{tabular}[c]{c}Scaling\\ (75\%)\end{tabular} & \begin{tabular}[c]{c}Gaussian Blur\\ ($\sigma$ = 0.1)\end{tabular} & \begin{tabular}[c]{@{}c@{}}Crop\\ (40\%)\end{tabular}  & \begin{tabular}[c]{@{}c@{}}JPEG Compression\\ (50\% quality)\end{tabular} &\begin{tabular}[c]{c}Combined\\(Crop, Gaussian blur, JPEG)\end{tabular} \\ 
\midrule
StegaNeRF~\cite{li2023steganerf}+3D-GS~\cite{kerbl20233d} & 93.15  & 54.48 & 67.22  &  73.98 & 73.84 & 75.87 & 73.28 & 76.71 \\
WateRF~\cite{jang2024waterf}+3D-GS~\cite{kerbl20233d} & 93.42 & 77.99 & 81.64 & 84.50 & 87.21 & 84.49 & 81.88 &  64.87  \\ \midrule
3D-GSW without FGD &  92.64  & 80.42 & 68.66 & 84.81 &  78.91 & 76.97 & 82.71 & 84.67 \\
3D-GSW (Ours) &  \textbf{97.37}  & \textbf{83.84} & \textbf{87.94} & \textbf{94.64} &  \textbf{96.01} & \textbf{92.86} & \textbf{92.65} & \textbf{90.84} \\
\bottomrule
\end{tabular}
}
\end{adjustbox}
\vspace{-.7em}
\caption{Quantitative evaluation of robustness under various attacks compared to baseline methods. The results are the average of Blender, LLFF, and Mip-NeRF 360 datasets. We conduct experiments using 32-bit messages. The best performances are highlighted in \textbf{bold}.}
\label{tab:robustness}
\vspace{-1.3em}
\end{table*}

\subsection{Experimental results}

\noindent \textbf{Rendering Quality and Bit Accuracy.} In this section, we compare the rendering quality and bit accuracy with other methods. As shown in Fig.~\ref{fig:figure3}, our method is most similar to the original and achieves high bit accuracy and rendering quality. In particular, since real-world scenes have complex structures, it is difficult to render them similarly to the original. From Fig.~\ref{fig:figure3}, while other methods have difficulty balancing the rendering quality and bit accuracy, our method achieves a good balance. Tab.~\ref{tab:invisibility_capacity} shows that our method ensures rendering quality and bit accuracy across all datasets compared to other methods. 
\\
\vspace{-.8em}

\noindent \textbf{Capacity of Message.}
Since bit accuracy, rendering quality, and capacity have a trade-off relationship. We explore this with message bit lengths $\{32, 48,64\}$. As shown in Tab.~\ref{tab:invisibility_capacity}, the bit accuracy, and rendering quality show a consistent decline as the message length increases. However, our method maintains a good balance between the invisibility and capacity of the message and outperforms the other methods as the message length becomes longer. Additionally, there is a further difference in performance compared to without FGD, depending on the message length. This shows that FGD is effective for large message embedding. From Fig.~\ref{fig:figure_length}, our method guarantees a good balance between bit accuracy and rendering quality. 
\\
\vspace{-0.8em}

\begin{table}[ht!]
\vspace{-.5em}
\setlength{\tabcolsep}{1pt}
\begin{adjustbox}{max width=\textwidth,center}
\centering
\scriptsize{
\begin{tabular}{@{}l|cccc@{}}
\toprule
&  \multicolumn{3}{c}{Bit Accuracy(\%) $\uparrow$}  \\
    \cmidrule{2-5}
    Methods & \begin{tabular}[c]{@{}c@{}}No \\ Distortion \end{tabular} & \begin{tabular}[c]{@{}c@{}} Adding \\ Gaussian Noise \\ ($\sigma$ = 0.1) \end{tabular} & \begin{tabular}[c]{@{}c@{}} Removing \\ 3D Gaussians \\ (20 \%) \end{tabular} & \begin{tabular}[c]{@{}c@{}} Cloning \\ 3D Gaussians \\ (20 \%) \end{tabular} \\ \midrule
    StegaNeRF~\cite{li2023steganerf}+3D-GS~\cite{kerbl20233d} &  93.15  & 61.82 & 60.24 & 75.56 \\
    WateRF~\cite{jang2024waterf}+3D-GS~\cite{kerbl20233d} & 93.42 & 73.85 & 80.58  & 82.32 \\ \midrule
    3D-GSW without FGD & 92.64 & 73.20 & 87.99 & 87.27 \\
    3D-GSW (Ours) & \textbf{97.37} & \textbf{89.11} & \textbf{96.87} & \textbf{95.99} \\
\bottomrule
\end{tabular}}
\end{adjustbox}
\vspace{-0.7em}
\caption{Robustness under model distortion. We show the results on 32-bits. The best performances are highlighted in \textbf{bold}.}
\label{tab:robustness_3d}
\vspace{-1.5em}
\end{table}
\noindent \textbf{Robustness for the image distortion.}
This section assesses the robustness of our method in situations where the rendered images are subjected to post-processing, which potentially modifies the embedded message within the rendered image. We evaluate the bit accuracy of the rendered images containing the message under various distortions. Tab.~\ref{tab:robustness} shows that other methods cannot guarantee robustness. In particular, the steganography method is weak to all attacks. Additionally, 3D-GSW without FGD, which does not remove 3D Gaussians, does not fully address robustness when embedding messages into the rendered image. In contrast, our method ensures robustness against all distortions by removing 3D Gaussians that interfere with robustness.
\\
\vspace{-.8em}

\noindent \textbf{Robustness for the 3D-GS distortion.} Since the purpose of our method is to protect both the rendered image and the core model, it is essential to consider the potential scenario of direct manipulation of the core model in cases of unauthorized model usage. To manipulate 3D-GS, we select to directly add Gaussian noise to the 3D-GS parameters. Additionally, we randomly remove and clone 3D Gaussians. As shown in Tab.~\ref{tab:robustness_3d}, our method is robust against 3D-GS distortion, outperforming the other methods. Furthermore, FGD robustly embeds the message into the rendered image, even if there is distortion in the model.
\begin{figure}[h!]
    \begin{center}
        \includegraphics[width=0.47\textwidth ]{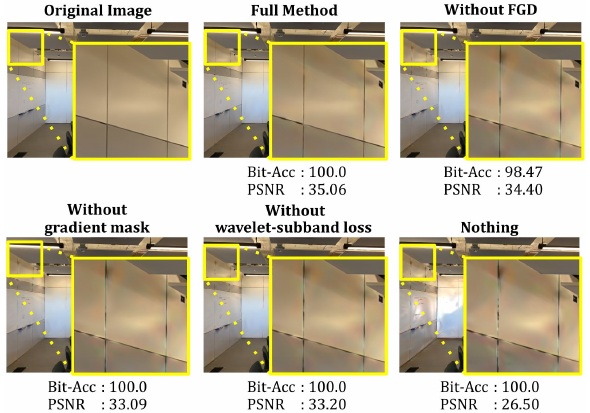}
    \end{center}
    \vspace{-1.7em}
    \caption{Rendering quality comparisons with full method(ours), without FGD, without gradient mask, without wavelet-subband loss, and base model. All images have 32-bits embedded.}
    \label{fig:figure5}
    \vspace{-1.5em}
\end{figure}

\begin{table}[ht!]
\vspace{-.5em}
\setlength{\tabcolsep}{5pt}
\centering
\scriptsize{
\begin{tabular}{@{}ccc|cccccccc@{}}
\toprule
\multicolumn{3}{c}{Methods} &  \multicolumn{4}{c}{Ours (3D-GSW)} & \\
  \cmidrule{1-7}
    FGD & Mask & $\mathcal{L}_{wavelet}$  & Bit Acc(\%)↑ & PSNR ↑ & SSIM ↑ & LPIPS ↓\\ \midrule
    -- & -- & -- & 96.50  & 29.96 & 0.951 & 0.072  \\
    $\checkmark$ & $\checkmark$ & -- & 96.16  & 33.56 & 0.971 & 0.052   \\
    $\checkmark$ & --  & $\checkmark$ & 96.37 & 33.26 & 0.967 & 0.054  \\
    --     & $\checkmark$ & $\checkmark$ & 94.60 & 34.27 & 0.975 & 0.047 \\
    $\checkmark$ & $\checkmark$ & $\checkmark$ & \textbf{97.37} &  \textbf{35.08} & \textbf{0.978}  &  \textbf{0.043}\\
\bottomrule
\end{tabular}
}
\vspace{-1em}
\caption{Quantitative ablation study of 3D-GSW shows that the best results are achieved when all components are combined. Results are shown for 32-bit messages.}
\label{tab:ablation_method}
\vspace{-2em}
\end{table}

\begin{figure}[ht!]
    \begin{center}
        \includegraphics[width=0.45\textwidth ]{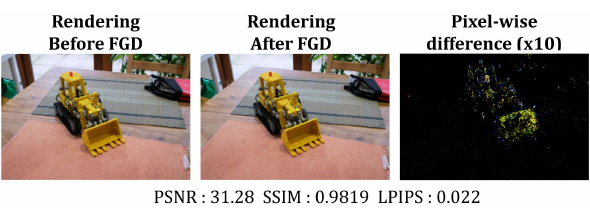}
    \end{center}
    \vspace{-2em}
    \caption{Qualitative result of  applying FGD.  We analyze the effect of FGD on the rendered image. Through FGD, we effectively control 3D Gaussians in the high-frequency area.}
    \vspace{-1em}
    \label{fig:figure7}
\end{figure}
\subsection{Ablation study} 

\noindent \textbf{FGD, Gradient mask, and Wavelet-subband loss.} In this section, we evaluate the effectiveness of FGD, gradient mask, and wavelet-subband loss. We remove each component in our method and compare the rendering quality with the bit accuracy. Fig.~\ref{fig:figure5} and Tab.~\ref{tab:ablation_method} show the results when each component is removed. First, we remove the FGD module in our method. In this case, our method tends to slightly decrease bit accuracy. Fig~\ref{fig:figure7} shows that FGD effectively adjusts 3D Gaussians in high-frequency areas, resulting in a quality that is nearly identical to the original. Second, without the gradient mask and wavelet-subband loss, our method performs poorly in preserving rendering quality. When all components are absent, our method fails to maintain an appropriate trade-off between bit accuracy and rendering quality, leading to a significant decrease in rendering quality. These results show the importance of each component in achieving a good balance between the rendering quality and bit accuracy.
\\
\vspace{-0.5em}




\begin{table}[h!]
\centering
\setlength{\tabcolsep}{10pt}
\scriptsize{
\begin{tabular}{@{}l|clll@{}}
\toprule
Subband              & Bit Acc↑ & PSNR ↑ & SSIM ↑ & LPIPS ↓ \\ \midrule
$LL, LH, HL, HH$    & 96.01  & 34.93  &  0.977 & 0.048   \\
$LH, HL, HH$ &  \textbf{97.37} & \textbf{35.08}  & \textbf{0.978}  & \textbf{0.043} 
\\\bottomrule
\end{tabular}
}
\vspace{-1.5em}
\caption{\label{tab:DWT_subband_loss} Ablation study on subband selection for wavelet-subband loss. Results represent the average score across Blender, LLFF, and Mip-NeRF 360 datasets using 32-bit messages.}
\vspace{-1em}
\end{table}
\noindent \textbf{Wavelet-subband loss.} 
Increasing the performance of both bit accuracy and rendering quality is challenging. To address this challenge, we design wavelet-subband loss. Since we modify 3D Gaussians in high-frequency areas, we utilize only the high-frequency subbands $\{LH, HL, HH\}$ to further ensure the rendering quality of those areas. Tab.~\ref{tab:ablation_method} and Fig.~\ref{fig:figure5} show that wavelet-subband loss effectively enhances rendering quality. Additionally, Tab.~\ref{tab:DWT_subband_loss} shows that using only high-frequency subbands results in higher rendering quality, with high bit accuracy.
\\
\vspace{-0.5em}

\begin{figure}[t!]
    \begin{center}
        \includegraphics[width=0.4\textwidth ]{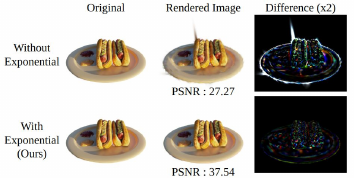}
    \end{center}
    \vspace{-1.7em}
    \caption{Qualitative comparison of the proposed gradient mask effect. For objects without a background, our method effectively adjusts 3D Gaussian parameters to prevent rendering beyond the object's boundary, preserving the original quality.}
    \vspace{-1.5em}
    \label{fig:figure8}
\end{figure}

\noindent \textbf{Gradient mask for 3D-GS.} Before the fine-tuning process, the pre-trained 3D-GS already has a high rendering quality. Since this property, if there is a large change in 3D-GS parameters, rendering quality can be decreased. When a gradient is propagated to a parameter, the gradient mask of our method ensures that the transmitted gradient is smaller than that of previous methods. Our gradient mask controls gradient transmission and minimizes parameter changes, thereby preserving rendering quality. Fig~\ref{fig:figure8} shows that our gradient mask (with exponential) enhances rendering quality more effectively than a previous method (without exponential). 
\\
\vspace{-.5em}

\begin{figure}[b!]
    \vspace{-2em}
    \begin{center}
        \includegraphics[width=0.47\textwidth ]{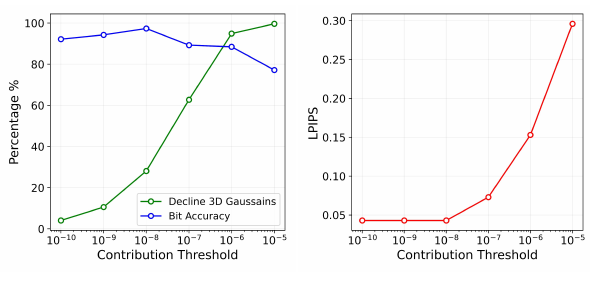}
    \end{center}
    \vspace{-3em}
    \caption{The impact of 3D Gaussians removal is based on the contribution of rendering quality. Declining 3D Gaussians refers to reducing the number of 3D Gaussians. The results are shown for 32-bit messages.}
    \vspace{-1.5em}
    \label{fig:figure6}
\end{figure}

\noindent \textbf{Control the number of 3D Gaussians.}
In this section, we present more details about the effect of controlling the number of 3D Gaussians. In the first phase of Frequency Guided Densification (FGD), we derive the contribution of rendering quality, $V_{\pi}$, for each 3D Gaussian. Fig.~\ref{fig:figure6} shows that removing 3D Gaussians with the contribution below $10^{-8}$ (removing 28.13 \%) has minimal impact on rendering quality and increases slightly bit accuracy. However, when FGD removal exceeds 50 \%, the bit accuracy and performance of LPIPS decrease. From the experimental results, reducing approximately 28\% 3D Gaussians preserves high bit accuracy and rendering quality.
\begin{table}[ht!]
\centering
\setlength{\tabcolsep}{7pt}
\footnotesize{
\begin{tabular}{@{}l|clll@{}}
\toprule
Methods              & Fine-tune ↓ & FPS ↑ & Storage ↓ \\ \midrule
3D-GS~\cite{kerbl20233d}  & -  & 56.65  & 833.89 MB  \\
StegaNeRF~\cite{li2023steganerf}+3D-GS~\cite{kerbl20233d}  & 58h 56m  & 56.65  &  833.89 MB  \\
WateRF~\cite{jang2024waterf}+3D-GS~\cite{kerbl20233d} & 6h 47m  & 56.65 & 833.89 MB \\ \midrule
3D-GSW (Ours) & \textbf{21m 03s} & \textbf{72.68} & \textbf{640.21 MB} 
\\\bottomrule
\end{tabular}
}
\vspace{-0.5em}
\caption{\label{tab:FPS} Results on the large-scale Mip-NeRF 360 dataset for 64-bits. All scores are averaged across Mip-NeRF 360 scenes, with 'fine-tunes' referring to embedded messages. For a fair comparison, we utilize the pre-trained models provided by 3D-GS~\cite{kerbl20233d}. The first row is the performance of the pre-trained models.}
\end{table}
\\

\noindent \textbf{Comparison of time and storage.} The advantages of 3D-GS are the high rendering quality and real-time rendering. However, the pre-trained 3D-GS contains redundant 3D Gaussians to achieve high-quality results, leading to storage capacity issues and increasing the time required for message embedding. Furthermore, other methods render twice during the fine-tuning process, resulting in inefficient embedding time for 3D-GS. To address these issues, we remove 3D Gaussians without a decrease in the rendering quality. Tab.~\ref{tab:FPS} shows that our method reduces storage of 3D-GS and message embedding time. 
In particular, our method enhances the real-time rendering. Notably, since the other methods maintain the number of 3D Gaussians, they follow the FPS and storage of pre-trained 3D-GS~\cite{kerbl20233d}.


\section{Conclusion}
We introduce the robust watermarking method for 3D Gaussian Splatting (3D-GS), developing a novel densification method, Frequency-Guided Densification (FGD), which ensures real-time rendering speed and robustness while improving rendering quality. We propose the gradient mask to ensure high rendering quality and introduce a wavelet-subband loss to enhance the rendering quality of high-frequency areas. Our experiments show that our method ensures the message and is robust against the distortion of the model compared to the other methods. Our method provides a strong foundation for exploring the broader implications and challenges of 3D-GS watermarking. It underscores the potential of advanced watermarking techniques to address ownership and security issues in the context of a rapidly evolving 3D industry.
In future work, we aim to extend our approach to embed multi-modal data, further broadening its applications and enhancing its utility in diverse domains. This expansion will broaden the scope of our method’s applications and enhance its adaptability and utility across a wide range of domains.
\\

\noindent \textbf{Limitations.}
Since our proposed method requires the pre-trained decoder, the decoder pre-training must be done first. Fortunately, the decoder only needs to be trained once per length of bits, and after training, the pre-training process for the corresponding length is not required. 
\vspace{5em}
\section*{Acknowledgements} {
This research was supported by 
the Culture, Sports and Tourism R\&D Program through the Korea Creative Content Agency grant funded by the Ministry of Culture, Sports and Tourism in 2024~( Research on neural watermark technology for copyright protection of generative AI 3D content, RS-2024-00348469, 47\%; International Collaborative Research and Global Talent Development for the Development of Copyright Management and Protection Technologies for Generative AI, RS-2024-00345025, 42\%),
the National Research Foundation of Korea(NRF) grant funded by the Korea government(MSIT)(RS-2025-00521602, 10\%), 
Institute of Information \& communications Technology Planning \& Evaluation (IITP) grant funded by the Korea government(MSIT) (No. RS-2019-II190079, Artificial Intelligence Graduate School Program(Korea University), 1\%), and
Artificial intelligence industrial convergence cluster development project funded by the Ministry of Science and ICT(MSIT, Korea)\&Gwangju Metropolitan City.
}
{
    \small
    \bibliographystyle{ieeenat_fullname}
    \bibliography{main}
}


\title{3D-GSW: 3D Gaussian Splatting for Robust Watermarking}


\clearpage
\setcounter{page}{1}
\maketitlesupplementary

\section*{Overview}
This supplementary material is organized as follows: 

Section~\hyperref[sup:1]{1} explains the implementation details of our watermark decoder. Section~\hyperref[sup:2]{2} elaborates on the additional details of Frequency-Guided Densification~(FGD). Section~\hyperref[sup:3]{3} provides the further analysis of the gradient mask. Section~\hyperref[sup:4]{4} presents the details of the wavelet-subband loss. Section~\hyperref[sup:5]{5} shows further experimental results, including the comparison with the baselines, the challenging scenario, the robustness test, the failure case, and qualitative results of all message bits 32, 48, and 64.

\section{Pre-training Decoder}
\label{sup:1}
In digital watermarking, the primary goal is to ensure that the message is decoded only from images where the message is embedded. In the fine-tuning method, a challenge arises: the model is typically trained with a unique message. This often leads to the decoder overfitting to a specific message, decoding the message from non-embedded rendered images. (See Tab.~\ref{tab:gt accuracy})

\begin{table}[htb!]
\vspace{-0.7em}
\setlength{\tabcolsep}{8pt}
\footnotesize{
\begin{tabular}{@{}l|clll@{}}
\toprule
Method              & Bit Acc↑~(W/ M) & Bit Acc↓~(W/O M)\\ \midrule
Train decoder with 3D-GS    & \textbf{99.53}  & 99.53     \\
Ours (pre-train decoder) &  97.37 & \textbf{53.92} 
\\\bottomrule
\end{tabular}
\vspace{-1em}
\caption{\label{tab:gt accuracy} Bit Acc↑(W/ M) is the bit accuracy from images with a message. Bit Acc↑(W/O M) is the bit accuracy from images without a message. The results are conducted on Blender, LLFF, and Mip-NeRF 360 datasets with watermarked message 32bits.}}
\end{table}
\vspace{-.5em}

To address the issue of the decoder memorizing the unique message, StegaNeRF~\cite{li2023steganerf} incorporates a classifier that determines the presence of a message. Furthermore, StegaNeRF~\cite{li2023steganerf} employs the specific patterns in the difference between the original image and the image containing the message, ensuring that the decoder cannot memorize the message.
However, this method requires prior knowledge of the specific type of attack applied to the image to decode the message. Since identifying the specific type of distortion used by unauthorized users is challenging, there is a need for a method capable of decoding messages directly from a single image, independent of the applied attack.
In our approach, we utilize a pre-trained decoder to prevent it from memorizing the unique message and to decode the message from a single image, ensuring robustness and reliability in the message decoding process.

\subsection{Architecture and Implementation Details}
\noindent{\textbf{Architecture.}} We utilize the HiDDeN architecture~\cite{zhu2018hidden}, which consists of the encoder, decoder, and distortion layer. The encoder has 4 Conv-BN-ReLU blocks, kernel size 3, stride 1, and padding 1. The decoder consists of 7 Conv-BN-ReLU blocks with 64 filters. 
Average pooling is performed over all spatial dimensions, and a final $L \times L$ linear layer produces the decoded message, where $L$ is the number of bits in the watermark message. The distortion layer has crop, scaling, and JPEG compression.
\\
\vspace{-.5em}

\noindent{\textbf{Implementation Details.}} We train the full HiDDeN architecture~\cite{zhu2018hidden} on \{32,48,64\} bits using the MS-COCO dataset~\cite{lin2014microsoft}. This process is only required once per bit length. The optimization is performed with Adam~\cite{kingma2014adam} and a learning rate of $10^{-4}$. We employ Mean Squared Error as an image loss and a message loss. To focus on decoder performance, we set the image loss hyper-parameter to 0 and the message loss hyper-parameter to 1 in the pre-training decoder process. By following the strategies from Stable Signature~\cite{fernandez2023stable}, we incorporate PCA whitening into the decoder to avoid bias in the decoded message, enhancing the robustness and reliability of the decoder.

\begin{table}[htb!]
\vspace{-1em}
\setlength{\tabcolsep}{4pt}
\centering
\footnotesize{
\begin{tabular}{@{}l|clll@{}}
\toprule
Input of Decoder  & Bit Acc(\%)↑ & PSNR ↑ & SSIM ↑ & LPIPS ↓ \\ \midrule
Pixel   &   70.93  &  31.63 &  0.973 & 0.050    \\
DFT & 50.10 & \textbf{40.36}  & 0.993  &  0.024  \\ 
DCT  &  50.08 &  39.97 & \textbf{0.994}  & \textbf{0.016}   \\   \midrule

LL Subband (Level 1)          & 94.17  & 30.86  & 0.972  & 0.051 \\ 
LL Subband (Level 2)          & \textbf{97.37}  & 35.08  & 0.978  & 0.043   \\
LL Subband (Level 3)          & 94.95  &  33.19 & 0.973 & 0.047   \\
LL Subband (Level 4)              & 90.06  & 32.41  & 0.970  &  0.021  \\\bottomrule

\end{tabular}
}
\vspace{-1em}
\caption{We transform the rendered image to the frequency domain, using DFT, DCT, and DWT. We choose low-frequency components for each transformed image as an input of the decoder to fine-tune 3D-GS. Pixel indicates that the rendered image itself is the input to the decoder. Results represent the average score across Blender, LLFF, and Mip-NeRF 360 datasets using 32-bit messages.}
\label{tab:table2}
\end{table}

\begin{table*}[h!]
\centering
\setlength{\tabcolsep}{5pt}
\renewcommand{\arraystretch}{0.5}
\begin{tabular}{@{}l|cccccr@{}}
\toprule
\multicolumn{1}{c|}{Dataset}  & \# Original & \# Remove & \# Split & \# After FGD & Removal Ratio & FPS↑~(Before/After/Ratio) \\ \midrule
Blender    & 0.29M  & 0.07M  &  0.03M & 0.25M  & 13.58\% & 208.05 / 231.62 /+11.33\% \\
LLFF &  0.98M  & 0.36M  &  0.06M & 0.68M  & 30.56\% & 75.24 / \ \ 90.55 /+20.35\% \\
Mip-NeRF 360 &  3.36M & 1.06M  & 0.09M  & 2.39M & 28.85\% & 56.65 / \ \ 72.68 /+28.30\% \\ \bottomrule
\end{tabular}
\caption{\label{tab:split_num} The impact of Frequency-Guided Densification (FGD) on the number of 3D Gaussians and rendering speed. We show the original number of 3D Gaussians, the number of removed and split 3D Gaussians, and the final count after FGD. The results are shown separately for Blender, LLFF, and Mip-NeRF 360 datasets.}
\end{table*}

\subsection{Select the input of Decoder} 
JPEG compression methods~\cite{liu2018deepn, sun2020reduction} tend to remove high-frequency of the image. WateRF~\cite{jang2024waterf} shows that the low-frequency of Discrete Wavelet Transform~(DWT) enables robust message embedding for radiance field watermarking. Following these properties, we choose the low-frequency as the input of the decoder. As shown in Tab.~\ref{tab:table2}, we observe that DWT at level 2, there is a good balance between bit accuracy and rendering quality. Notably, other transformations like Discrete Fourier Transform~(DFT) and Discrete Cosine Transform~(DCT) fail to embed the message. Furthermore, the pixel domain can embed messages but does not guarantee high bit accuracy. From these results, we choose DWT at level 2 to embed the message into 3D-GS.

\begin{table}[t!]
\centering
\setlength{\tabcolsep}{2.5pt}
\small{
\begin{tabular}{@{}l|clll@{}}
\toprule
Methods              & Bit Acc↑ & PSNR ↑ & SSIM ↑ & LPIPS ↓ \\ \midrule
3D-GSW (W/O split)    &  96.72 & 33.72  &  0.970 & 0.062   \\
3D-GSW (W/ split) (Ours) &  \textbf{97.37} & \textbf{35.08}  & \textbf{0.978}  & \textbf{0.043} 
\\\bottomrule
\end{tabular}
}
\vspace{-1em}
\caption{\label{tab:No_split} We compare bit accuracy and rendering quality of FGD with split~(W/ split) and without split~(W/O split). Results represent the average score across Blender, LLFF, and Mip-NeRF 360 datasets using 32-bit messages.}
\vspace{-1em}
\end{table}

\begin{figure}[t!]
    \begin{center}
        \includegraphics[width=0.45\textwidth ]{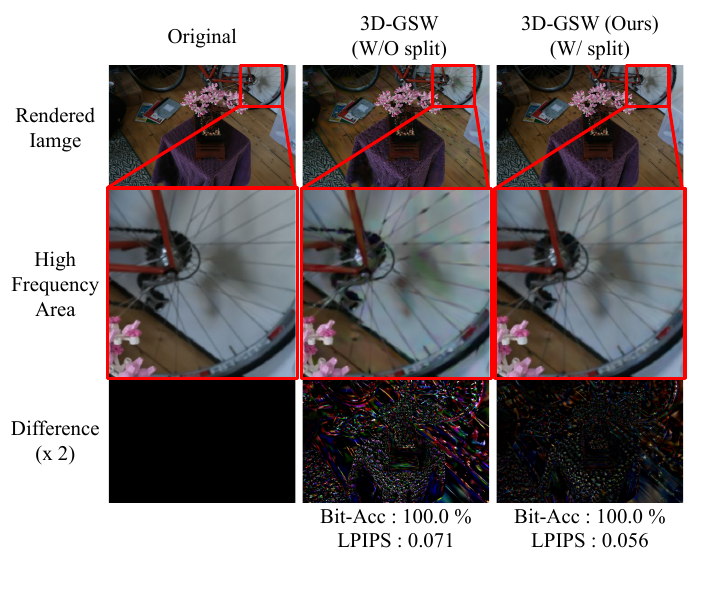}
    \end{center}
    \vspace{-3em}
    \caption{We compare the difference between FGD with split~(W/ split) and without split~(W/O split). Both images are embedded by 32-bit messages.}
    \vspace{-1em}
    \label{fig:compare_qualitive_split}
\end{figure}

\section{Frequency-Guided Densification}
\label{sup:2}
\subsection{Effectiveness of Split 3D Gaussians}
From 3D-GS compression work~\cite{lee2023compact}, small 3D Gaussians have a negligible contribution to the rendering quality due to their minimal volume. As noted in image compression work~\cite{liu2018deepn}, human visual system is less sensitive to high-frequency components. Furthermore, as highlighted by error-based densification~\cite{bulo2024revising}, since a small number of large 3D Gaussians are responsible for capturing high-frequency details, there are cases where 3D-GS has difficulty reconstructing high-frequency areas. To address these properties, we choose 3D Gaussians in high-frequency areas to preserve the rendering quality of the pre-trained 3D-GS. Additionally, to effectively render high-frequency details during the fine-tuning process, each selected 3D Gaussian is split into two smaller 3D Gaussians.

As shown in Tab.~\ref{tab:No_split}, while both methods ensure high bit accuracy, our FGD significantly enhances the rendering quality. From Fig.~\ref{fig:compare_qualitive_split}, splitting 3D Gaussians reconstructs the high-frequency components in more detail. Notably, FGD, which does not split 3D Gaussians, covers the high-frequency component by increasing the volume of 3D Gaussians.
These results quantitatively and qualitatively demonstrate the effectiveness and efficiency of split 3D Gaussians. Furthermore, as shown in Tab.~\ref{tab:split_num}, although splitting increases the number of 3D Gaussians, the number of removed 3D Gaussians is relatively higher. Consequently, the total number of 3D Gaussians after FGD is reduced, leading to improved rendering efficiency while maintaining quality. 

\begin{table}[t!]
\centering
\setlength{\tabcolsep}{2.5pt}
\footnotesize{
\begin{tabular}{@{}l|clll@{}}
\toprule
Methods              & Bit Acc↑ & PSNR ↑ & SSIM ↑ & LPIPS ↓ \\ \midrule
3D-GSW (FGD W/ DWT)  & 95.12   & 34.23 &  0.976 &   0.054    \\
3D-GSW (FGD W/ DCT)  & 95.48   & 34.20  & 0.976  & 0.054   \\ \midrule
3D-GSW (FGD W/ DFT) (Ours) &  \textbf{97.37} & \textbf{35.08}  & \textbf{0.978}  & \textbf{0.043} 
\\\bottomrule
\end{tabular}
}
\vspace{-1em}
\caption{\label{tab:other_frequency_patch} We compare the effectiveness of DFT with other methods for transforming patches into the frequency domain. Results represent the average score across Blender, LLFF, and Mip-NeRF 360 datasets using 32-bit messages.}
\vspace{-1.5em}
\end{table}

\begin{figure}[t!]
    \begin{center}
        \includegraphics[width=0.45\textwidth ]{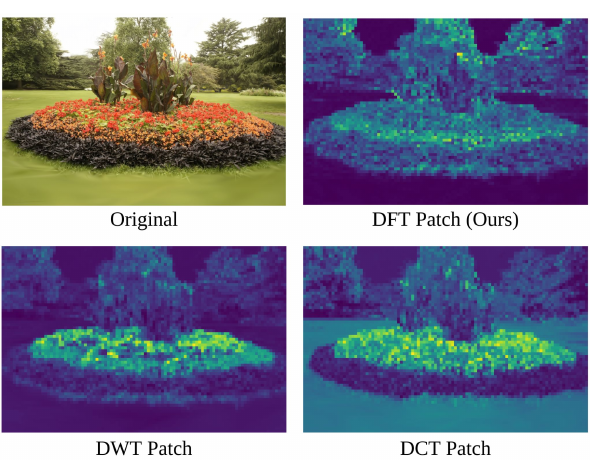}
    \end{center}
    \vspace{-2em}
    \caption{We visualize the patches converted by DFT, DWT, and DCT, applying a mask that emphasizes high frequency.}
    \vspace{-1em}
    \label{fig:compare_patch_frequency}
\end{figure}

\begin{figure*}[t!]
    \begin{center}
        \includegraphics[width=1\textwidth ]{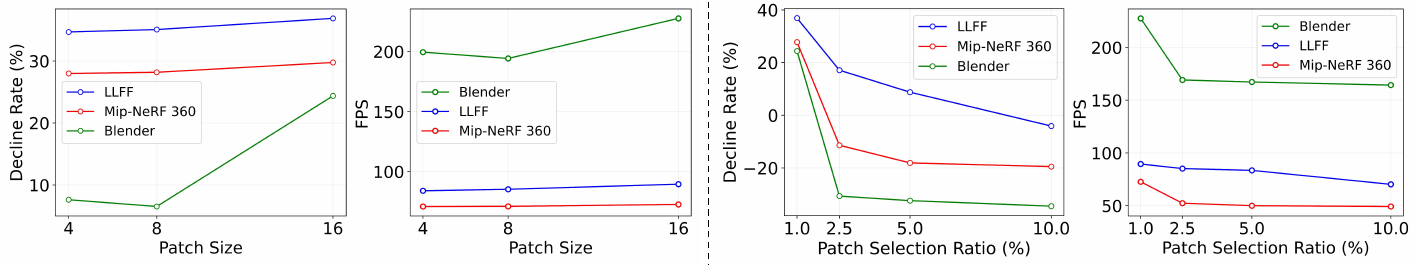}
    \end{center}
    \vspace{-2em}
    \caption{We compare the effects of patch size and patch selection ratio on the number of 3D Gaussians in 3D-GS after undergoing FGD. For patch size, we set patch selection ratio to 1 \%. For patch selection ratio, we set patch size to 16 $\times$ 16.}
    \vspace{-1em}
    \label{fig:compare_patch_proportion}
\end{figure*}

\subsection{Emphasize High-Frequency Patch}
In phase 2 of FGD, we find the patches with strong intensity of high-frequency signals to split 3D Gaussians within those patches.
To measure the intensity of high-frequency signals globally across the patch, Discrete Fourier Transform (DFT) is applied. By analyzing the frequency signals of the patch through DFT, the value of the high-frequency areas is calculated to quantify their intensity through a mask to emphasize high-frequency signals in the patch~(See main paper Sec.3.4). 

To evaluate the effectiveness of DFT, we also utilize other frequency transforms, such as DWT and DCT. 
In the case of DWT, we decompose the patch into wavelet subbands and measure the intensity as the average value of the high-frequency subbands. For DCT, we transform the patch into the frequency domain and measure the intensity as the average value of the high-frequency signals in the DCT.

As shown in Tab.~\ref{tab:other_frequency_patch}, our method enhances bit accuracy and rendering quality, which have a trade-off relationship. Furthermore, from Fig.~\ref{fig:compare_patch_frequency}, we observe that DWT detects high-frequency regions in specific image areas, leveraging its ability to analyze local information. In contrast, DCT has difficulty in detecting high-frequency areas. 
Notably, DFT effectively detects high-frequency areas across the entire image due to its characteristic of analyzing global information.
These results show that DFT effectively evaluates the distribution of high-frequency signals within the patch.

\begin{table}[t!]
\centering
\setlength{\tabcolsep}{8.5pt}
\footnotesize{
\begin{tabular}{@{}l|cllll@{}}
\toprule
Patch Size              & Bit Acc↑ & PSNR ↑ & SSIM ↑ & LPIPS ↓  \\ \midrule
4 $\times$ 4 & 94.12   & 33.13 &  0.975 &   0.054    \\
8 $\times$ 8 & 95.48   & 34.20  & 0.976  & 0.050   \\ 
16 $\times$ 16 (Ours) &  \textbf{97.37} & \textbf{35.08}  & \textbf{0.978}  & \textbf{0.043}
\\\bottomrule
\end{tabular}
}
\vspace{-1em}
\caption{\label{tab:patch_size} We compare performance by patch size. Results represent the average score across Blender, LLFF, and Mip-NeRF 360 datasets using 32-bit messages.}
\vspace{-1em}
\end{table}

\begin{table}[t!]
\centering
\setlength{\tabcolsep}{9.5pt}
\footnotesize{
\begin{tabular}{@{}l|cllll@{}}
\toprule
Top K \%              & Bit Acc↑ & PSNR ↑ & SSIM ↑ & LPIPS ↓  \\ \midrule
10  &  94.21  & 32.34  & 0.975  & 0.061 \\
5 &  95.49  & 33.06 & 0.975  & 0.050      \\
2.5  &  97.19  & 33.37  & \textbf{0.978}  & 0.044   \\ 
1 (Ours) &  \textbf{97.37} & \textbf{35.08}  & \textbf{0.978}  & \textbf{0.043} 
\\\bottomrule
\end{tabular}
}
\vspace{-1em}
\caption{\label{tab:selection_K} We compare performance by selection ratio, denoted as K. Results represent the average score across Blender, LLFF, and Mip-NeRF 360 datasets using 32-bit messages.}
\vspace{-1em}
\end{table}

\subsection{Patch Size and Selection Patch}
In this section, we explore how patch size and patch selection ratio affect performance. 
Since we remove 3D Gaussians to embed a robust message in phase 1 of FGD, it is crucial to avoid splitting more Gaussians than are removed during phase 2 of FGD. 
As shown in Fig.~\ref{fig:compare_patch_proportion}, we observe that increasing the patch size split 3D Gaussians without disturbing the results of removal 3D Gaussians in phase 1, enhancing real-time rendering. Tab.~\ref{tab:patch_size} shows that a large patch size improves bit accuracy and rendering quality. Fig.~\ref{fig:compare_patch_proportion} shows that selecting more patches causes more 3D Gaussians to be split, increasing the number of 3D Gaussians after FGD. Additionally, Tab.~\ref{tab:selection_K} shows that as the patch selection ratio increases, all performance decreases. These results show that an appropriate patch size and patch selection are effective for message embedding and rendering quality.

\section{Gradient Mask}
\label{sup:3}
In this section, we present the additional analysis about a gradient mask. 
Our gradient mask is calculated as follows: 
\begin{equation}
\begin{aligned}
w = \frac{1}{e^{\vert \theta \vert^{\beta}}}, \quad  z = \frac{w}{\sum_{i=1}^{N_{\mathcal{G}^{\prime}_o}} w_i} 
\end{aligned}
\label{eq:adaptive mask}
\end{equation}
, where $\mathcal{G}^{\prime}_o$, $\theta$, $N_{\mathcal{G}^{\prime}_o}$, i and $\beta > 0$ are respectively denoted as 3D-GS passed through FGD, the parameter of $\mathcal{G}^{\prime}_o$, the number of 3D Gaussians in $\mathcal{G}^{\prime}_o$, the index of 3D Gaussians and the strength of gradient manipulation.

\subsection{Effectiveness of Exponential Function}
In FGD process, we modify 3D Gaussians in the pre-trained 3D-GS to enhance the robustness of the message and rendering quality.
Since 3D-GS passed through FGD provides high-quality rendering, it is crucial to preserve the rendering quality during the fine-tuning process. To achieve this, we reduce gradient size to minimize changes in the parameters of 3D-GS passed through FGD, utilizing an exponential function. As shown in Tab.~\ref{tab:exponential_similar}, we observe that the exponential function maintains the parameters of 3D-GS passed through FGD. Notably, the exponential function preserves color parameters identically. Tab.~\ref{tab:exponential_metric} shows that the exponential function has superior rendering quality compared to other methods.

\begin{table}[htb!]
\centering
\setlength{\tabcolsep}{3pt}
\footnotesize{
\begin{tabular}{@{}c|ccccc@{}}
\toprule
\multicolumn{1}{c|}{} &  \multicolumn{5}{c}{Cosine Similarity of Parameters ↑}  \\
  \cmidrule{2-6}
    Methods & Color & Opacity & Scale & Rotation & Position \\ \midrule
    W/O exponential & 0.982 & \textbf{0.997} & 0.991 & 0.989 & \textbf{0.999}  \\
    W/ exponential (Ours) & \textbf{0.999} & \textbf{0.997} & \textbf{0.999}  & \textbf{0.995} & \textbf{0.999} 
\\\bottomrule
\end{tabular}
}
\vspace{-1em}
\caption{\label{tab:exponential_similar} We compare similarity of parameters for a exponential function. We calculate cosine similarity between 3D-GS passed through fine-tuning process and 3D-GS passed through FGD process. Results represent the average score across Blender, LLFF, and Mip-NeRF 360 datasets using 32-bit messages.}
\end{table}

\begin{table}[htb!]
\centering
\setlength{\tabcolsep}{3pt}
\footnotesize{
\begin{tabular}{@{}l|cllll@{}}
\toprule
Methods              & Bit Acc↑ & PSNR ↑ & SSIM ↑ & LPIPS ↓  \\ \midrule
3D-GS After FGD (N/M)  &  - &  \textbf{48.66} & \textbf{0.998}  & \textbf{0.002} \\
StegaNeRF~\cite{li2023steganerf}+3D-GS~\cite{kerbl20233d} &  93.15 &   32.68 &  0.953 & 0.049 \\
WateRF~\cite{jang2024waterf}+3D-GS~\cite{kerbl20233d}  & 93.42 & 30.49 & 0.956 & 0.050 \\
W/O exponential   &  93.99 & 27.54  &  0.898 & 0.093  \\
W/ exponential (Ours) &  \textbf{97.37} & 35.08  & 0.978  & 0.043 
\\\bottomrule
\end{tabular}
}
\caption{\label{tab:exponential_metric} We compare bit accuracy and rendering quality for a exponential function with other methods. 3D-GS After FGD (N/ M) refers to 3D-GS that has gone through FGD before embedding the message. Results represent the average score across Blender, LLFF, and Mip-NeRF 360 datasets using 32-bit messages.}
\end{table}

\begin{figure}[t!]
    \begin{center}
        \includegraphics[width=0.47\textwidth ]{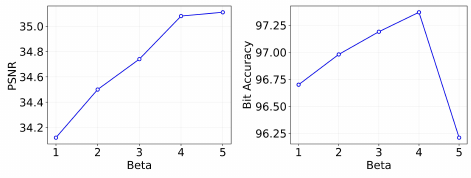}
    \end{center}
    \vspace{-1em}
    \caption{We compare the strength of the gradient mask. Beta controls the size of the gradient. A larger beta reduces the gradient size further.}
    \label{fig:compare_beta}
\end{figure}

\subsection{Strength of Gradient Mask}
Following Eq.~\ref{eq:adaptive mask}, $\beta$ controls the strength of gradient manipulation. As shown in Fig.~\ref{fig:compare_beta}, we observe that increasing $\beta$ enhances the rendering quality. However, bit accuracy does not continue to increase but decreases when $\beta$ becomes greater than 4. This shows the trade-off between bit accuracy and rendering quality, and we find that $\beta=4$ achieves a good balance between bit accuracy and rendering quality.

\section{Wavelet-subband loss}
\label{sup:4}

\begin{figure}[htb!]
    \begin{center}
        \includegraphics[width=0.47\textwidth ]{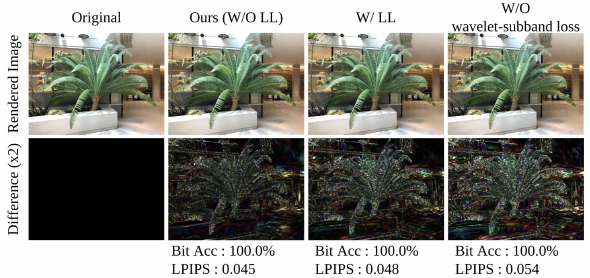}
    \end{center}
    \vspace{-1em}
    \caption{We show the differences(x2) between the watermarked image and the original image. Our method enhances the detail in the high-frequency areas.}
    \label{fig:supple_subband_loss_compare}
\end{figure}

\subsection{Effectiveness of Wavelet-subband loss}
Since 3D-GS increases the volume of fewer 3D Gaussians to render high-frequency areas, there is a tendency to lose detailed information in the high-frequency areas. To address this issue, in phase 2 of FGD, we split 3D Gaussians into smaller ones. Furthermore, we propose a wavelet-subband loss to avoid losing details in the high-frequency areas during the fine-tuning process. Wavelet-subband loss is designed to focus on the high-frequency, utilizing only high-frequency subbands (LH, HL, HH). 

\begin{figure}[htb!]
    \begin{center}
        \includegraphics[width=0.47\textwidth ]{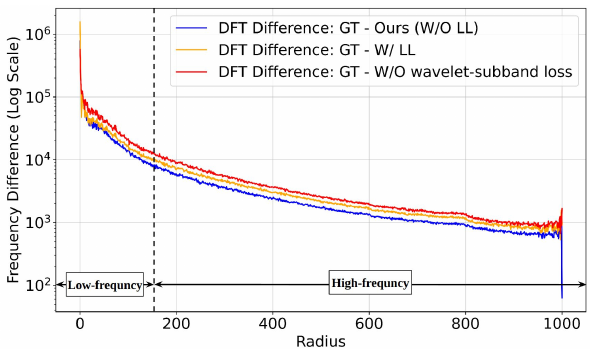}
    \end{center}
    \vspace{-1em}
    \caption{We show the differences in the frequency domain between the watermarked image and the original image. We transform both images to the frequency domain using DFT. Radius is the distance from the center of the image. The larger radius represents the high-frequency areas. The blue line represents our result. A lower value on the y-axis indicates greater similarity to the original image. We show this result on the LLFF dataset using 32-bit messages.}
    \label{fig:supple_subband_loss}
\end{figure}

As shown in Fig.~\ref{fig:supple_subband_loss_compare}, we observed that the wavelet-subband loss using only high-frequency subbands achieves better rendering quality. To further analyze the rendering quality in the high-frequency areas, we convert both the original image and the rendered image to the frequency domain and calculate the difference between them. Fig.~\ref{fig:supple_subband_loss} shows that our method is similar to using LL subbands in the low-frequency areas, but achieves results more similar to the original in the high-frequency areas.
These results quantitatively and qualitatively show the effectiveness of wavelet-subband loss in enhancing quality in high-frequency areas.

\section{Additional Results}
\label{sup:5}

\subsection{Comparison with the baselines}
\begin{figure}[htb!]
    \begin{center}
        \includegraphics[width=0.45\textwidth ]{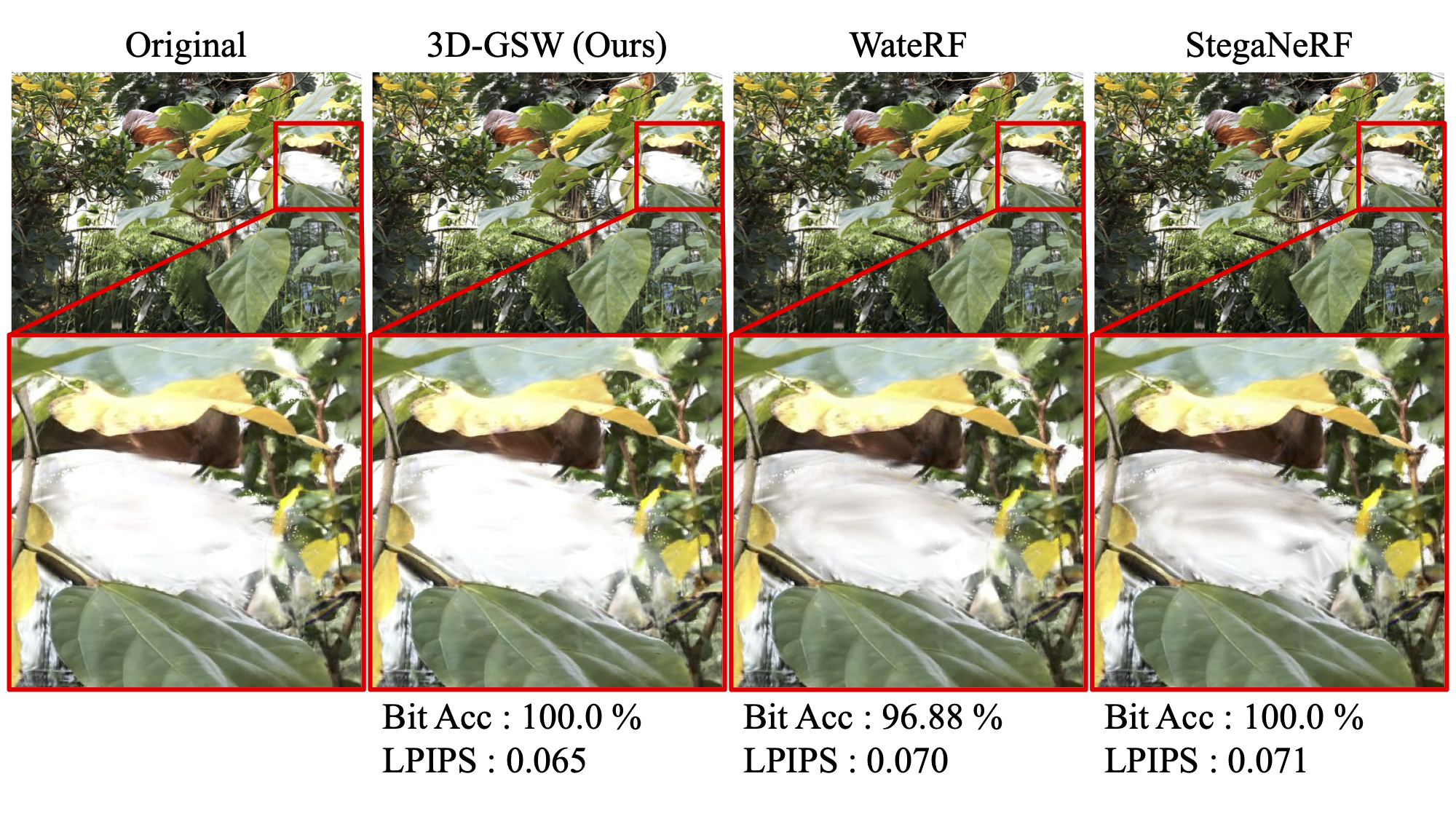}
    \end{center}
    \vspace{-2em}
    \caption{We compare the rendering quality with other methods. We achieve results that are most similar to the original rendering quality.}
    \label{fig:supple_compare}
\end{figure}

\begin{figure}[h!]
    \begin{center}
        \includegraphics[width=0.4\textwidth ]{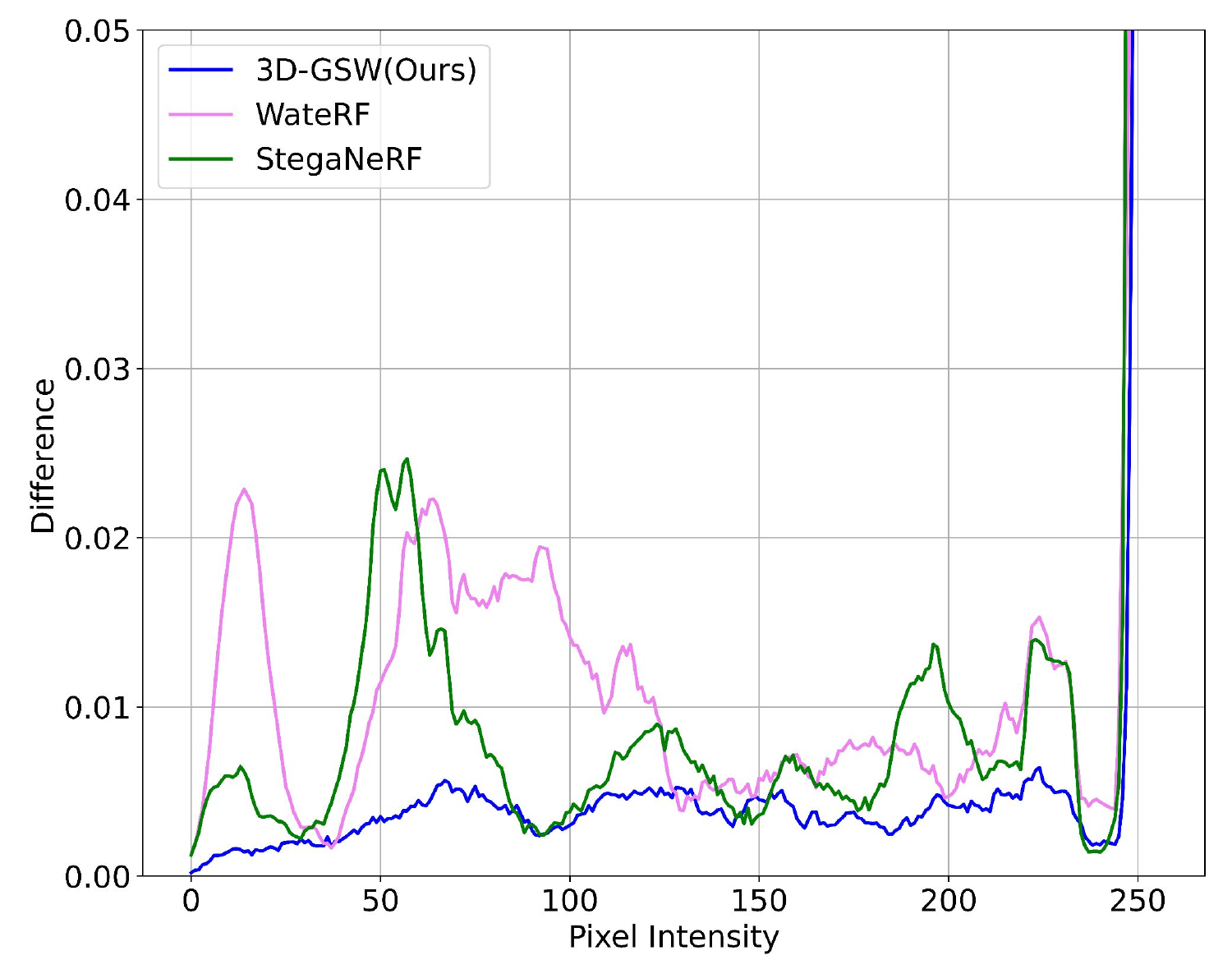}
    \end{center}
    \vspace{-2em}
    \caption{We show the normalized pixel intensity difference between original and the watermarked image. The blue line represents our result. A lower value on the y-axis indicates greater similarity to the original image.}
    \vspace{-1em}
    \label{fig:supple_pixel_intensity}
\end{figure}

As shown in Fig.~\ref{fig:supple_compare}, we show the rendering quality with other methods. We observe that WateRF~\cite{jang2024waterf} and StegaNeRF~\cite{li2023steganerf} have color artifacts in the rendered image due to increasing the volume of 3D Gaussians. In contrast, our method preserves rendering quality, while achieving high bit accuracy. To illustrate the differences between the original and watermarked image, we visualize the normalized pixel intensity difference in Fig.~\ref{fig:supple_pixel_intensity}, following previous watermarking methods~\cite{cai2017process,reed2015watermarking}. Our method achieves the smallest difference, demonstrating superior rendering quality.

\subsection{Robustness Test}
\begin{figure*}[t!]
    \begin{center}
        \includegraphics[width=0.9\textwidth ]{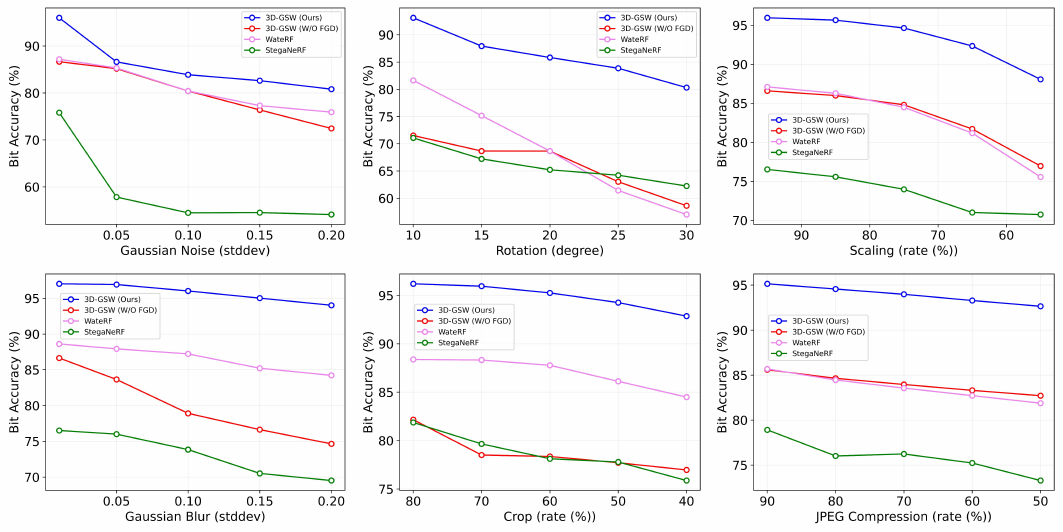}
    \end{center}
    \vspace{-2em}
    \caption{Bit accuracy for WateRF~\cite{jang2024waterf}, StegaNeRF~\cite{li2023steganerf}, 3D-GSW~(W/O FGD), and our method for various image distortions and distortion strengths. The blue line represents our results. Our method outperforms other methods.}
    \label{fig:supple_robustness}
\end{figure*}

\begin{figure*}[t!]
    \begin{center}
        \includegraphics[width=0.9\textwidth ]{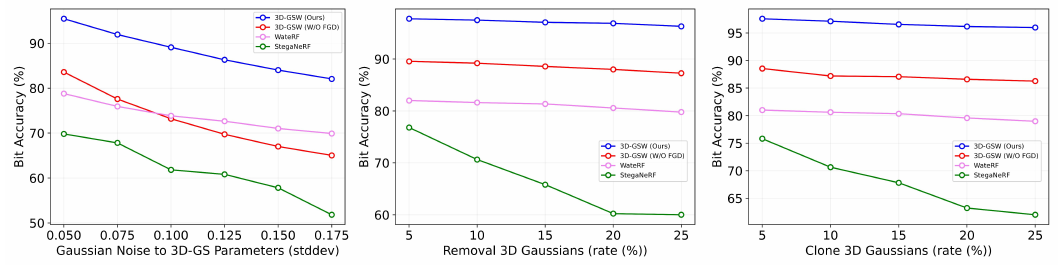}
    \end{center}
    \vspace{-2em}
    \caption{Bit accuracy for WateRF~\cite{jang2024waterf}, StegaNeRF~\cite{li2023steganerf}, 3D-GSW~(W/O FGD), and our method for model distortions and distortion strengths. We conduct two model distortion: 1) We add model Gaussian noise to 3D-GS parameters. 2) We remove randomly 3D Gaussians.  3) We clone randomly 3D Gaussians. The blue line represents our results. Our method outperforms other methods.}
    \label{fig:supple_robustness_model}
\end{figure*}

Fig.~\ref{fig:supple_robustness} and Fig.~\ref{fig:supple_robustness_model} show the bit accuracy when the distortion level is varied. We observe that our bit accuracy is consistently higher than other methods across a wide range of strengths. Notably, incorporating FGD improves performance further, highlighting its role in enhancing robustness to image and model distortions. These results show that our method is a highly effective and robust watermarking method.

\subsection{Challenging Scenario}

To demonstrate the practicality and scalability, we experiment on a 4D-GS task with a method named Spacetime~\cite{li2024spacetime} using the neural 3D Video dataset~\cite{li2022neural}, which consists of six dynamic scenes with a resolution of 2028×2704. 
Tab.~\ref{tab:4D case} shows a performance comparison between ours and baselines. 
Our method consistently achieves superior performance on dynamic scenes, particularly in terms of temporal consistency. This highlights its effectiveness in handling dynamic scenes while maintaining high rendering quality. Furthermore, the scalability and practicality of our method make it well-suited for various real-world applications.

\begin{table}[h!]
\centering
\setlength{\tabcolsep}{1.3pt}
\footnotesize{
\begin{tabular}{@{}l|cccccc@{}}
\toprule
\multicolumn{1}{c|}{Method}  & Bit-Acc↑ & PSNR↑ & SSIM↑ & LPIPS↓ & FVD↓ \\ \midrule
StegaNeRF~\cite{li2023steganerf}+Spacetime~\cite{li2024spacetime}       & 94.75\% & 31.26  & 0.940  & 0.0895 & 315.11   \\
WateRF~\cite{jang2024waterf}+Spacetime~\cite{li2024spacetime}          & 95.11\% & 30.51  & 0.939  & 0.0909 & 166.15  \\
3D-GSW~+Spacetime~\cite{li2024spacetime}  & \textbf{97.62\%} & \textbf{32.51}  & \textbf{0.958}  & \textbf{0.0653} & \textbf{125.98}  
\\\bottomrule
\end{tabular}
}
\vspace{-1em}
\caption{\label{tab:4D case} We show the scalability of our method to dynamic scenes. Results represent the average score across neural 3D Video dataset~\cite{li2022neural} using 32-bit messages.}
\end{table}

\subsection{Failure Case}

Recently, compression works have emerged in the spotlight in 3D-GS. Thus, we can consider one scenario, in which unauthorized users compress watermarked 3D-GS. To perform the compression of pre-trained 3D-GS, the unauthorized users must have the same training data of the target 3D asset to train the 3D-GS. We assume that the unauthorized users have the same training data as the watermarked 3D-GS and proceed with compression. The compression is conducted by Simon~\cite{niedermayr2024compressed}. As shown in Tab.~\ref{tab:failure case}, we observe that all methods lose the message in the compression process. This result is a failure case of the entire radiance field watermarking research. Although our method is robust when the unauthorized users only have 3D-GS, we will work on the case in the future when they proceed with compression.

\begin{table}[h!]
\centering
\setlength{\tabcolsep}{3pt}
\footnotesize{
\begin{tabular}{@{}l|cllll@{}}
\toprule
Methods              & Bit Acc↑ & PSNR ↑ & SSIM ↑ & LPIPS ↓  \\ \midrule
StegaNeRF~\cite{li2023steganerf}+3D-GS~\cite{kerbl20233d} &  52.29 & 38.68 &  \textbf{0.986} & \textbf{0.018} \\
WateRF~\cite{jang2024waterf}+3D-GS~\cite{kerbl20233d}  & 50.86 & 38.77 & 0.985 & \textbf{0.018} \\
3D-GSW (Ours) &  \textbf{55.51} & \textbf{38.83}  & \textbf{0.986}  & \textbf{0.018} 
\\\bottomrule
\end{tabular}
}
\vspace{-1em}
\caption{\label{tab:failure case} We show the impact of the compression on bit accuracy and rendering quality. All methods lose the embedded message due to compression.}
\end{table}

\subsection{Qualitative results}
\begin{figure*}[ht!]
    \begin{center}
        \includegraphics[width=0.9\textwidth ]{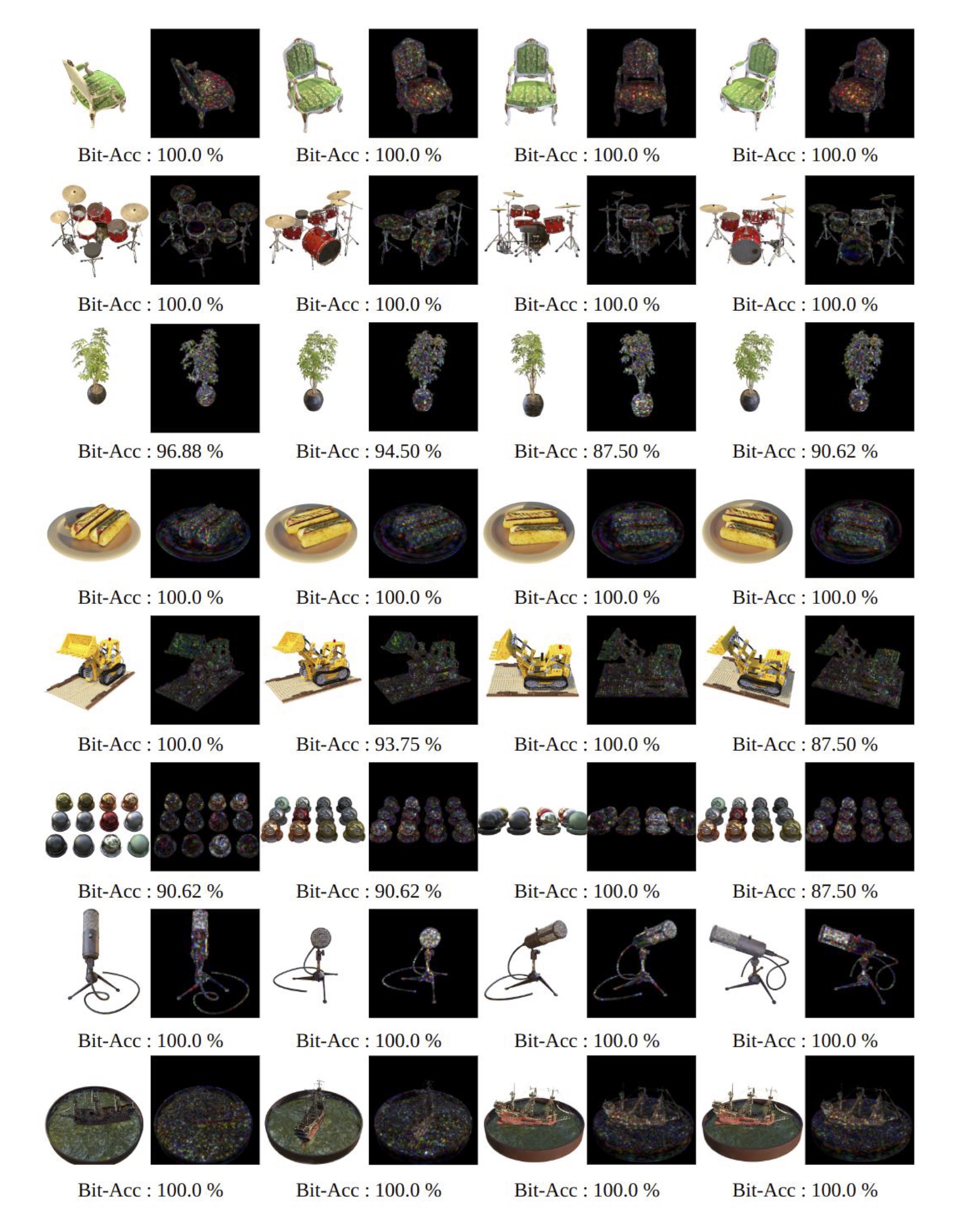}
    \end{center}
    \vspace{-2em}
    \caption{Rendering quality of various rendering outputs using our method on Blender dataset. We show the differences ($\times$ 2). The closer it is to white, the bigger the difference between the ground truth and the image. We show the results on 32 bits.}
    \vspace{-1em}
    \label{fig:supple_blender32}
\end{figure*}

\begin{figure*}[ht!]
    \begin{center}
        \includegraphics[width=0.9\textwidth ]{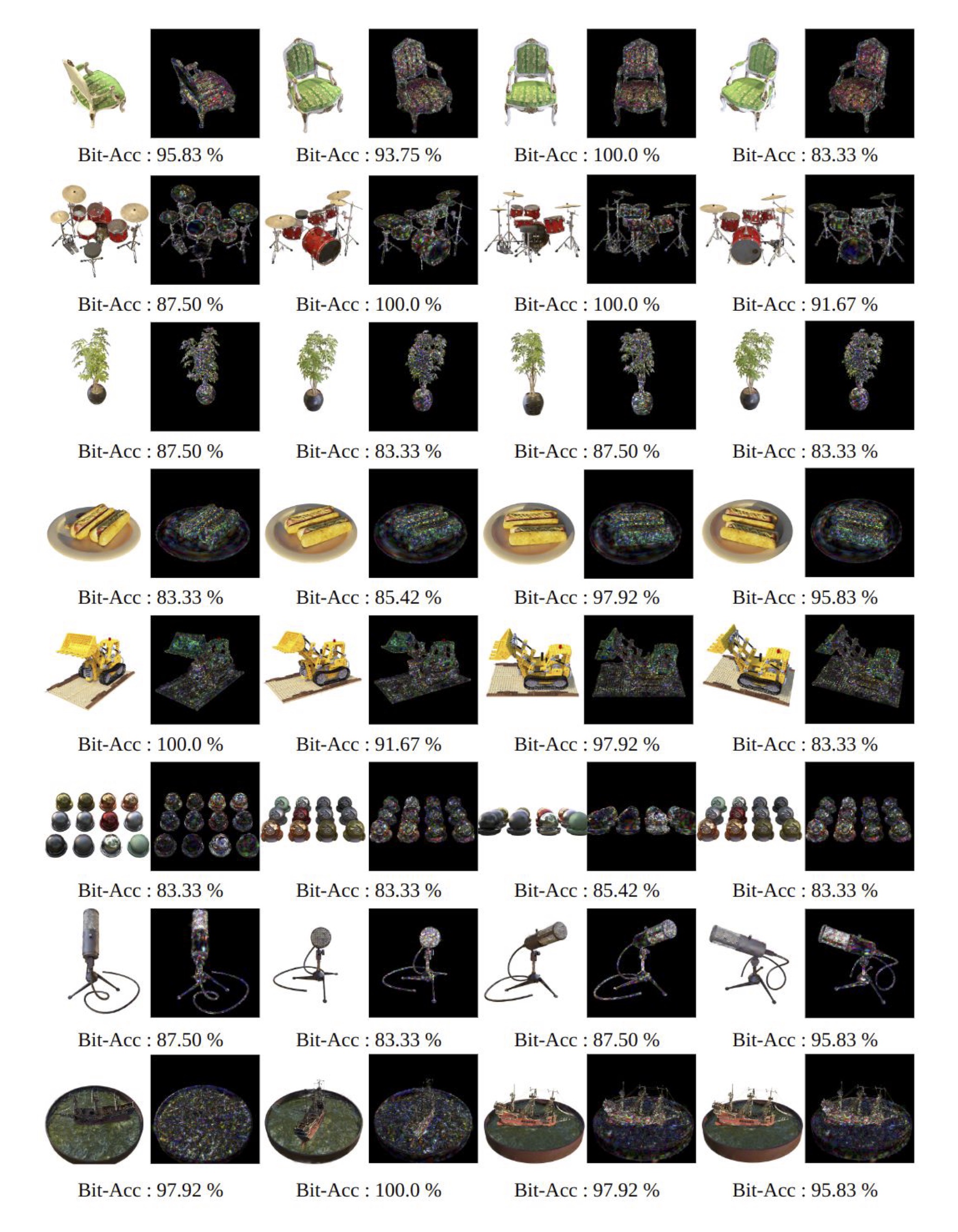}
    \end{center}
    \vspace{-2em}
    \caption{Rendering quality of various rendering outputs using our method on Blender dataset. We show the differences ($\times$ 2). The closer it is to white, the bigger the difference between the ground truth and the image. We show the results on 48 bits.}
    \vspace{-1em}
    \label{fig:supple_blender48}
\end{figure*}

\begin{figure*}[ht!]
    \begin{center}
        \includegraphics[width=0.9\textwidth ]{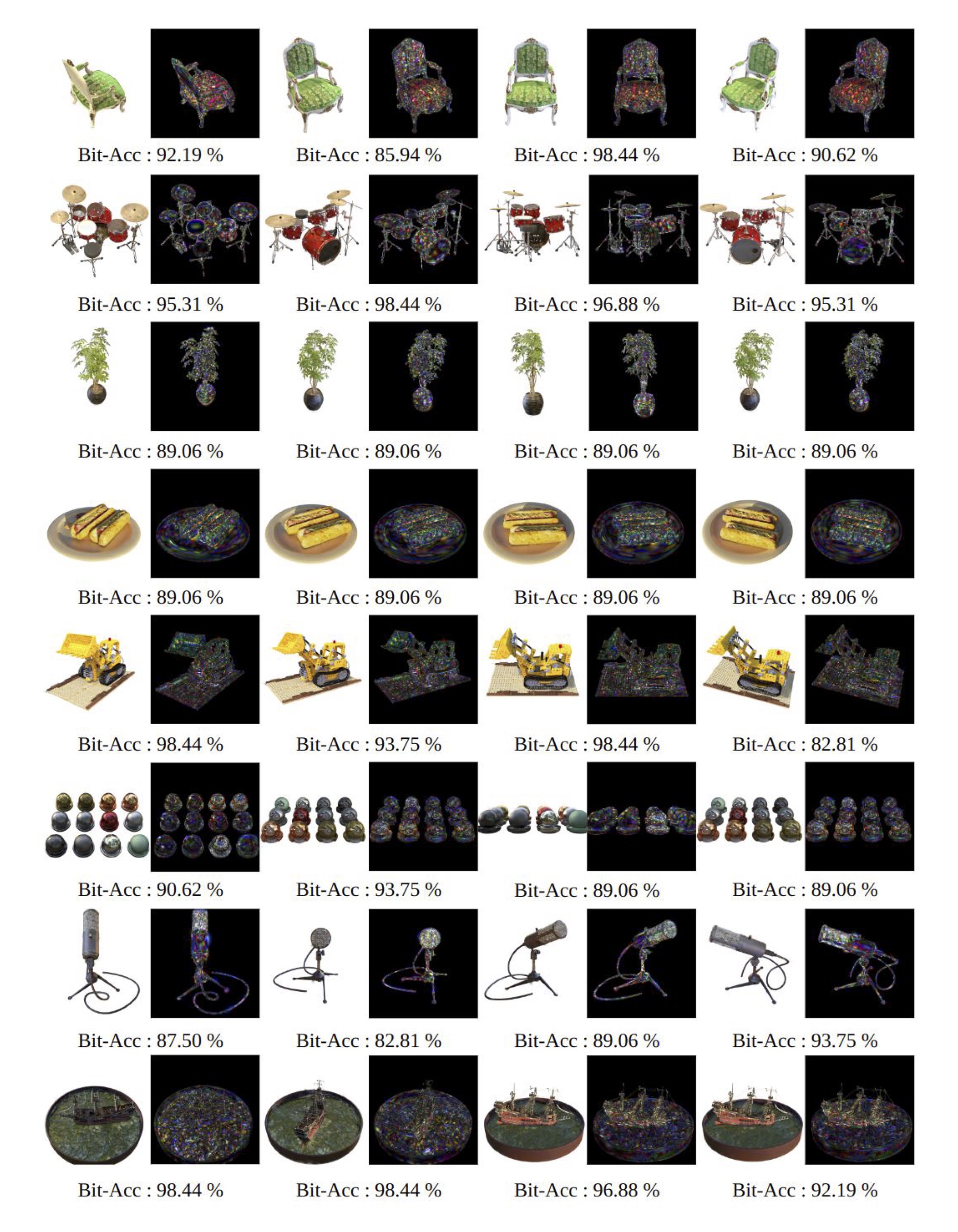}
    \end{center}
    \vspace{-2em}
    \caption{Rendering quality of various rendering outputs using our method on Blender dataset. We show the differences ($\times$ 2). The closer it is to white, the bigger the difference between the ground truth and the image. We show the results on 64 bits.}
    \vspace{-1em}
    \label{fig:supple_blender64}
\end{figure*}

\begin{figure*}[ht!]
    \begin{center}
        \includegraphics[width=1\textwidth ]{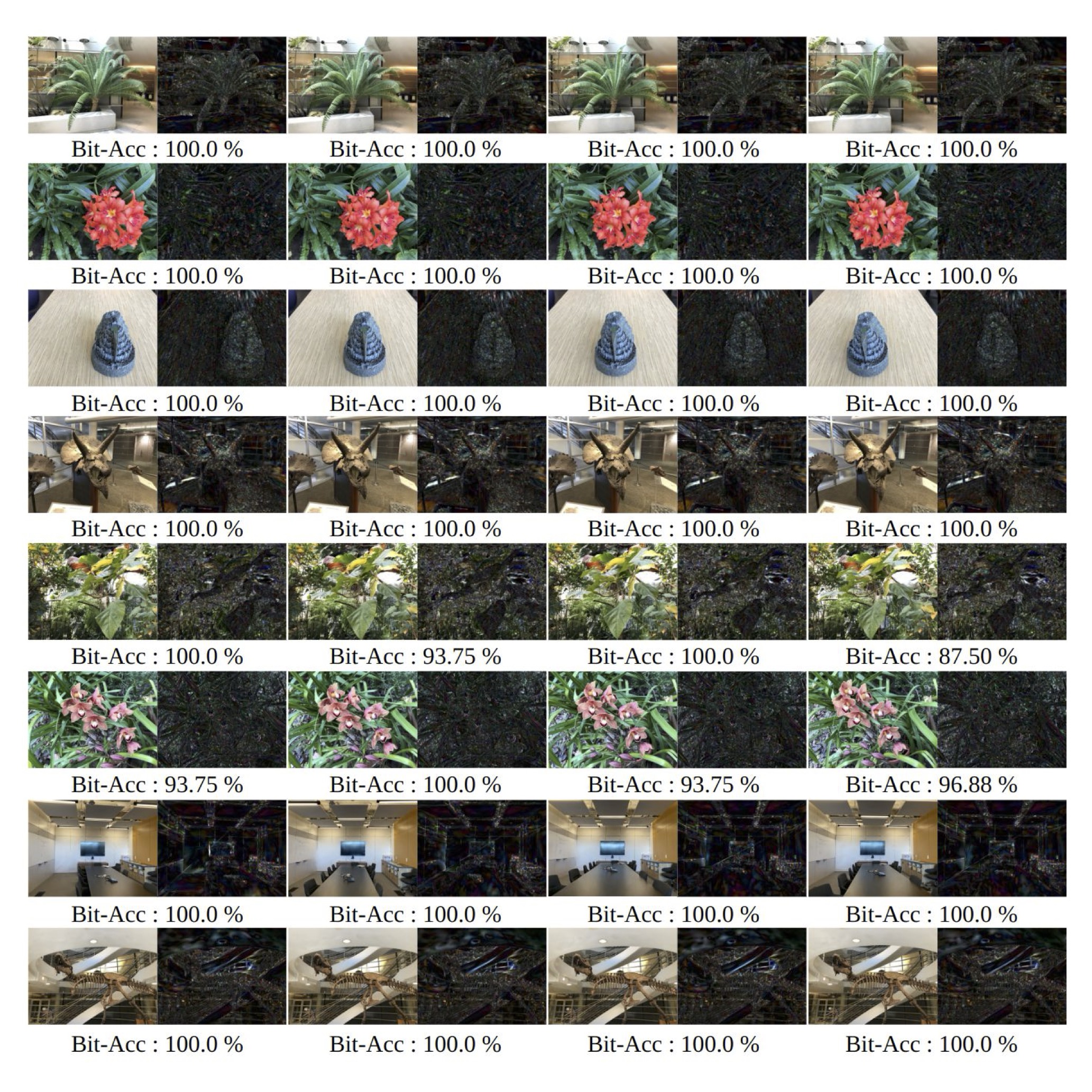}
    \end{center}
    \vspace{-2em}
    \caption{Rendering quality of various rendering outputs using our method on LLFF dataset. We show the differences ($\times$ 2). The closer it is to white, the bigger the difference between the ground truth and the image. We show the results on 32 bits.}
    \vspace{-1em}
    \label{fig:supple_llff32}
\end{figure*}

\begin{figure*}[ht!]
    \begin{center}
        \includegraphics[width=1\textwidth ]{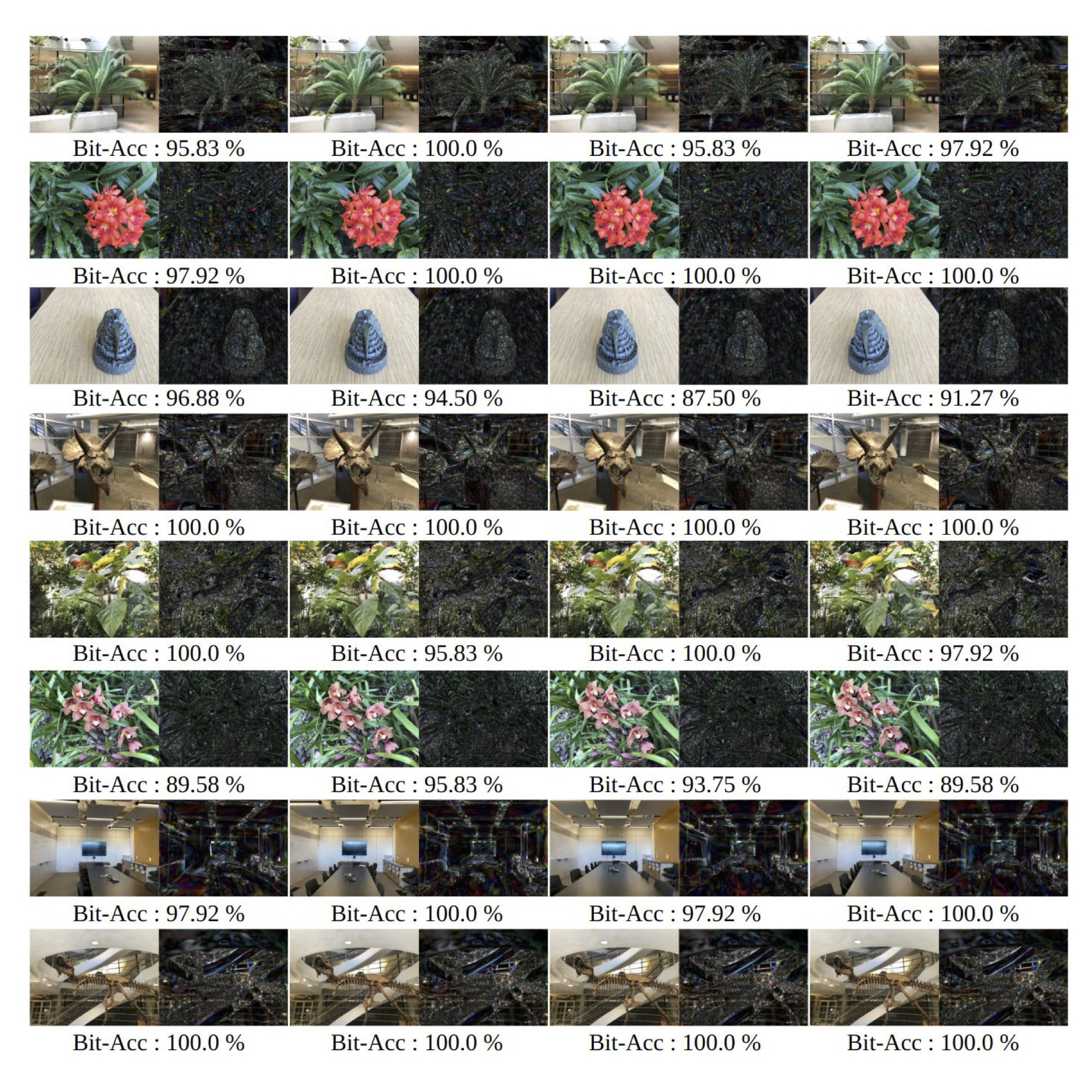}
    \end{center}
    \vspace{-2em}
    \caption{Rendering quality of various rendering outputs using our method on LLFF dataset. We show the differences ($\times$ 2). The closer it is to white, the bigger the difference between the ground truth and the image. We show the results on 48 bits.}
    \vspace{-1em}
    \label{fig:supple_llff48}
\end{figure*}

\begin{figure*}[ht!]
    \begin{center}
        \includegraphics[width=1\textwidth ]{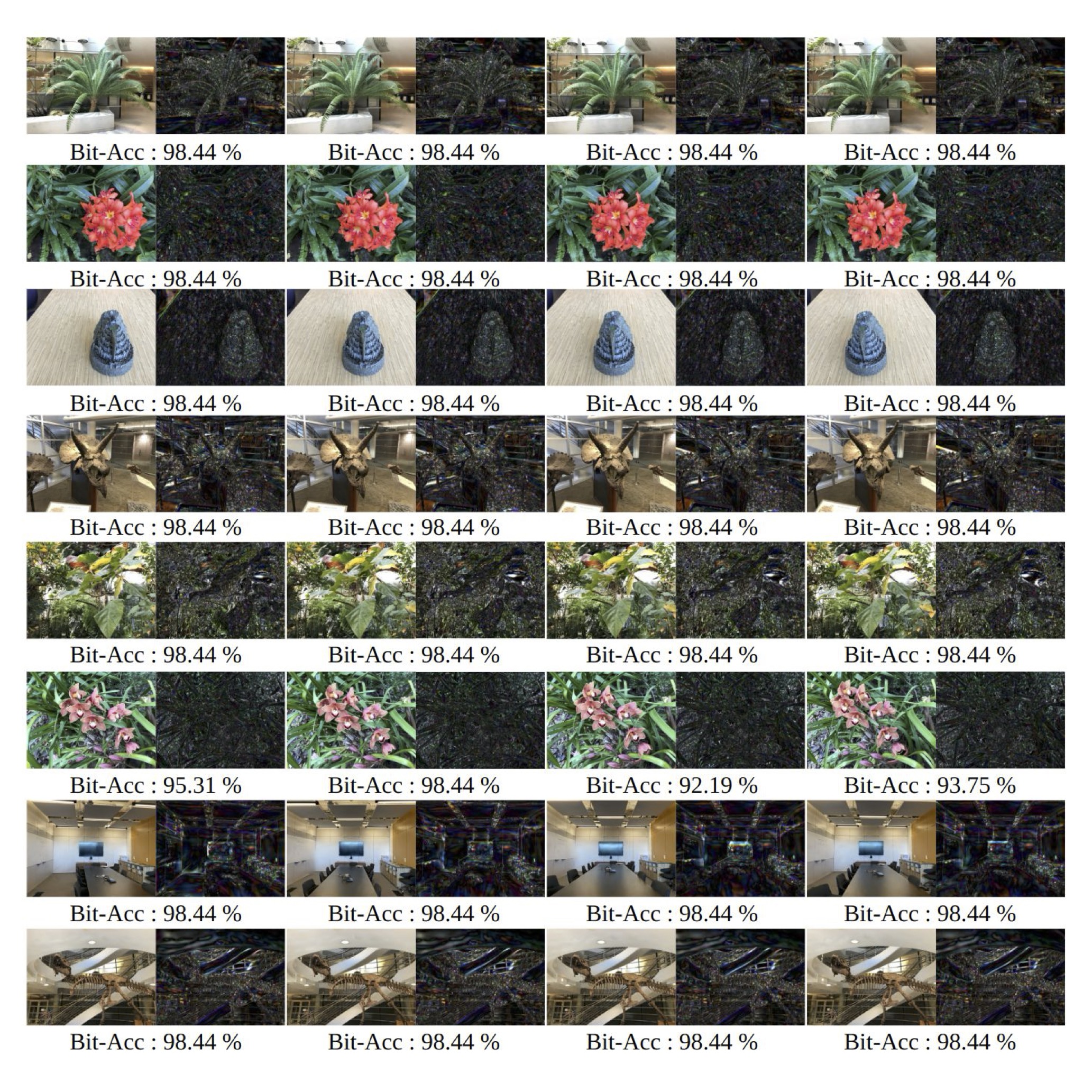}
    \end{center}
    \vspace{-2em}
    \caption{Rendering quality of various rendering outputs using our method on LLFF dataset. We show the differences ($\times$ 2). The closer it is to white, the bigger the difference between the ground truth and the image. We show the results on 64 bits.}
    \vspace{-1em}
    \label{fig:supple_llff64}
\end{figure*}

\begin{figure*}[ht!]
    \begin{center}
        \includegraphics[width=1\textwidth ]{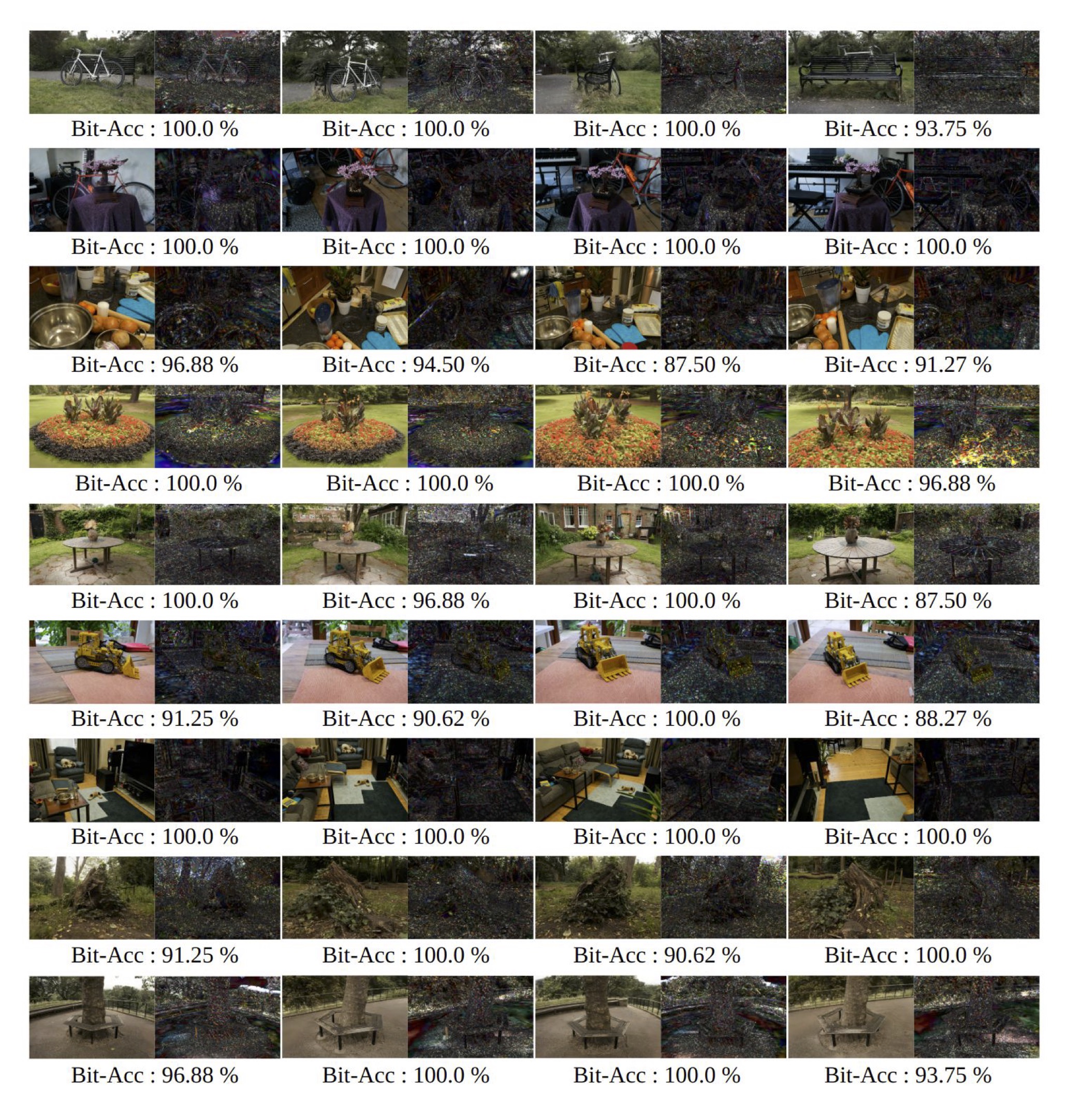}
    \end{center}
    \vspace{-2em}
    \caption{Rendering quality of various rendering outputs using our method on Mip-NeRF 360 dataset. We show the differences ($\times$ 2). The closer it is to white, the bigger the difference between the ground truth and the image. We show the results on 32 bits.}
    \vspace{-1em}
    \label{fig:supple_mip32}
\end{figure*}

\begin{figure*}[ht!]
    \begin{center}
        \includegraphics[width=1\textwidth ]{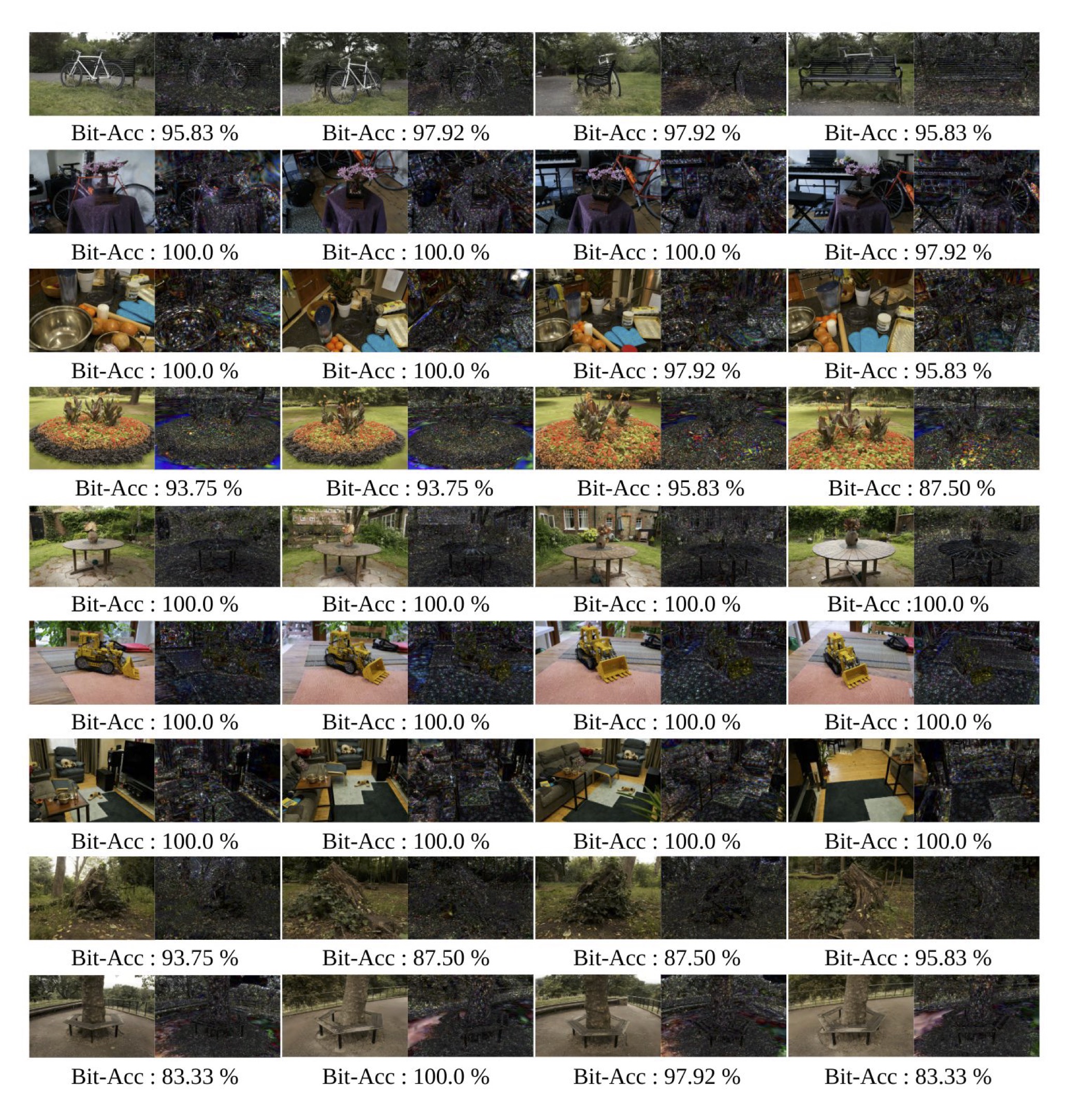}
    \end{center}
    \vspace{-2em}
    \caption{Rendering quality of various rendering outputs using our method on Mip-NeRF 360 dataset. We show the differences ($\times$ 2). The closer it is to white, the bigger the difference between the ground truth and the image. We show the results on 48 bits.}
    \vspace{-1em}
    \label{fig:supple_mip48}
\end{figure*}

\begin{figure*}[ht!]
    \begin{center}
        \includegraphics[width=1\textwidth ]{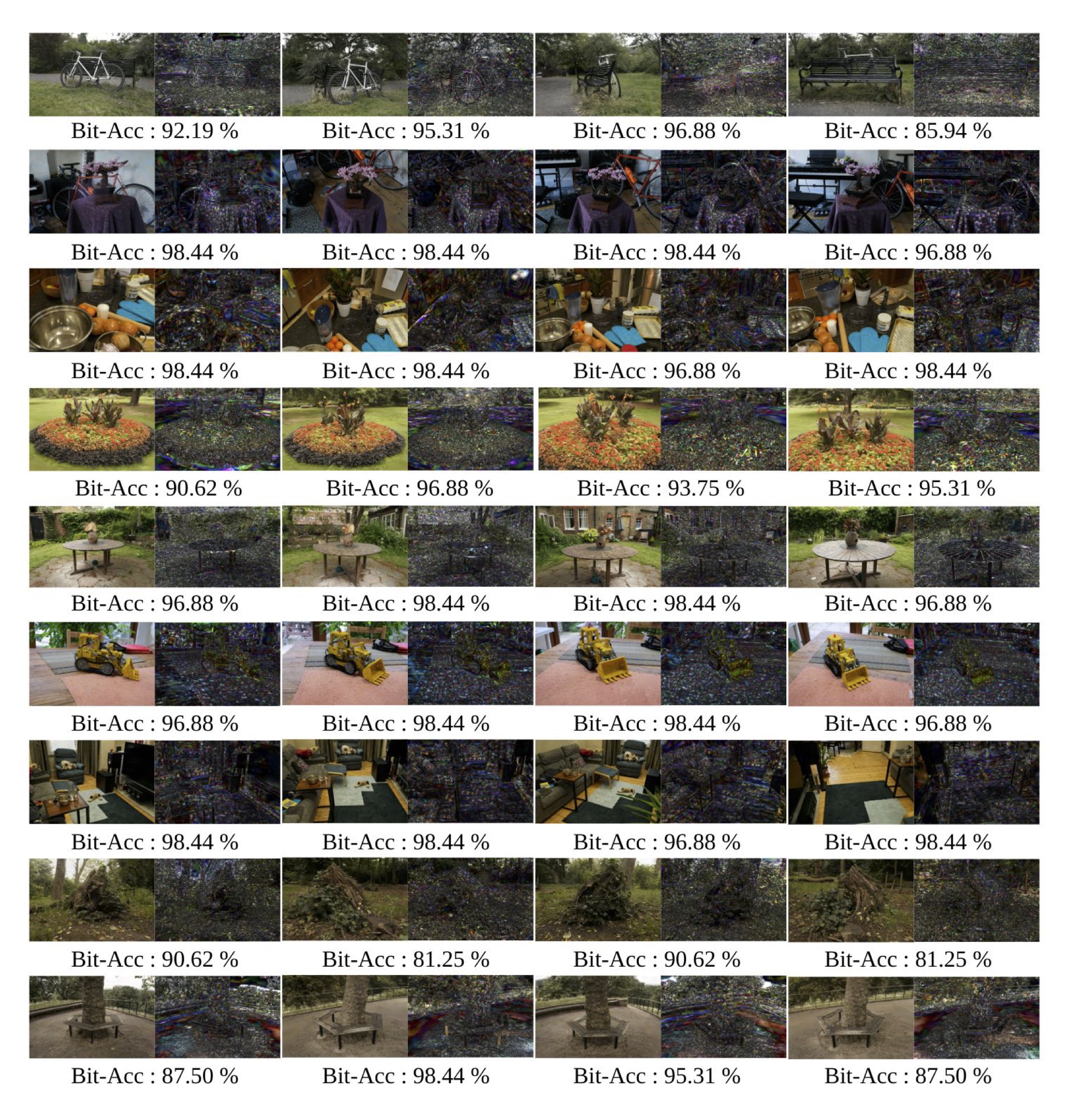}
    \end{center}
    \vspace{-2em}
    \caption{Rendering quality of various rendering outputs using our method on Mip-NeRF 360 dataset. We show the differences ($\times$ 2). The closer it is to white, the bigger the difference between the ground truth and the image. We show the results on 64 bits.}
    \vspace{-1em}
    \label{fig:supple_mip64}
\end{figure*}

From Fig.~\ref{fig:supple_blender32} to Fig.~\ref{fig:supple_mip64}, we visualize all results rendered from our method and the difference ($\times$ 2) between the original image and watermarked image.

\clearpage

\clearpage
{
    \small
}

\end{document}